\documentclass[10pt,twocolumn,letterpaper]{article}

\usepackage{iccv}
\usepackage{times}
\usepackage{epsfig}
\usepackage{graphicx}
\usepackage{subfigure} 
\usepackage{stackengine}
\usepackage{xcolor}
\usepackage{hhline}
\usepackage[english]{babel}
\usepackage{bm}
\usepackage{amsmath}
\usepackage{amssymb}
\usepackage[abs]{overpic}

\newcommand{\vpara}[1]{\vspace{0.1in}\noindent\textbf{#1}}

\usepackage[pagebackref=true,breaklinks=true,letterpaper=true,colorlinks,bookmarks=false]{hyperref}

 \iccvfinalcopy 

\def\iccvPaperID{180} 

\ificcvfinal\fi
\begin{document}

\title{Arbitrary Style Transfer in Real-time with Adaptive Instance Normalization}

%
\author{Xun Huang \qquad Serge Belongie \\
	{Department of Computer Science \& Cornell Tech, Cornell University}\\
	{\tt\small \{xh258,sjb344\}@cornell.edu}
}

\maketitle

\begin{abstract}
Gatys~et~al.~recently introduced a neural algorithm that renders a content image in the style of another image, achieving so-called style transfer. However, their framework requires a slow iterative optimization process, which limits its practical application. Fast approximations with  feed-forward neural networks have been proposed to speed up neural style transfer. Unfortunately, the speed improvement comes at a cost: the network is usually tied to a fixed set of styles and cannot adapt to arbitrary new styles. In this paper, we present a simple yet effective approach that for the first time enables arbitrary style transfer in real-time. At the heart of our method is a novel adaptive instance normalization~(AdaIN) layer that aligns the mean and variance of the content features with those of the style features. Our method achieves speed comparable to the fastest existing approach, without the restriction to a pre-defined set of styles. In addition, our approach allows flexible user controls such as content-style trade-off, style interpolation, color \& spatial controls, all using a single feed-forward neural network.
\end{abstract}

\vspace{-0.12in}
\section{Introduction}
The seminal work of Gatys \etal~\cite{gatys2016image} showed that deep neural networks~(DNNs) encode not only the content but also the \emph{style} information of an image. Moreover, the image style and content are somewhat separable: it is possible to change the style of an image while preserving its content. The style transfer method of \cite{gatys2016image} is flexible enough to combine content and style of arbitrary images. However, it relies on an optimization process that is  prohibitively slow. 

Significant effort has been devoted to accelerating neural style transfer. \cite{johnson2016perceptual, ulyanov2016texture, li2016precomputed} attempted to train feed-forward neural networks that perform stylization with a single forward pass. 
A major limitation of most feed-forward methods is that each network is restricted to a single style. There are some recent works addressing this problem, but they are either still limited to a finite set of styles~\cite{dumoulin2017learned,li2017diversified,zhang2017multi,chen2017stylebank}, 
or much slower than the single-style transfer methods~\cite{chen2017fast}. 

In this work, we present the first neural style transfer algorithm that resolves this fundamental flexibility-speed dilemma. Our approach can transfer arbitrary new styles in real-time, combining the flexibility of the optimization-based framework~\cite{gatys2016image} and the speed similar to the fastest feed-forward approaches~\cite{johnson2016perceptual,ulyanov2017improved}. 
Our method is inspired by the \emph{instance normalization}~(IN)~\cite{ulyanov2017improved,dumoulin2017learned} layer, which is surprisingly effective in feed-forward style transfer. To explain the success of instance normalization, we propose a new interpretation that instance normalization performs style normalization by normalizing feature statistics, which have been found to carry the style information of an image~\cite{gatys2016image,li2016combining,li2017demystifying}. Motivated by our interpretation, we introduce a simple extension to IN, namely \emph{adaptive instance normalization~}(AdaIN).
Given a content input and a style input, AdaIN simply adjusts the mean and variance of the content input to match those of the style input. Through experiments, we find AdaIN effectively combines the content of the former and the style latter by transferring feature statistics. A decoder network is then learned to generate the final stylized image by inverting the AdaIN output back to the image space. Our method is nearly three orders of magnitude faster than~\cite{gatys2016image}, without sacrificing the flexibility of transferring inputs to arbitrary new styles. Furthermore, our approach provides abundant user controls at runtime, without any modification to the training process.
\section{Related Work}
\vspace{-0.1in}
\vpara{Style transfer.}  The problem of style transfer has its origin from non-photo-realistic rendering~\cite{kyprianidis2013state}, and is closely related to texture synthesis and transfer~\cite{efros1999texture,efros2001image,elad2016style}. Some early approaches include histogram matching on linear filter responses~\cite{heeger1995pyramid} and non-parametric sampling~\cite{efros2001image,frigo2016split}. These methods typically rely on low-level statistics and often fail to capture semantic structures. Gatys \etal~\cite{gatys2016image} for the first time demonstrated impressive style transfer results by matching feature statistics in convolutional layers of a DNN. Recently, several improvements to~\cite{gatys2016image} have been proposed. Li and Wand~\cite{li2016combining} introduced a framework based on markov random field~(MRF) in the deep feature space to enforce local patterns. Gatys \etal~\cite{gatys2017controlling} proposed ways to control the color preservation, the spatial location, and the scale of style transfer. Ruder \etal~\cite{ruder2016artistic} improved the quality of video style transfer by imposing temporal constraints. 


The framework of Gatys~\etal~\cite{gatys2016image} is based on a slow optimization process that iteratively updates the image to minimize a content loss and a style loss computed by a loss network. It can take minutes to converge even with modern GPUs. On-device processing in mobile applications is therefore too slow to be practical. 
A common workaround is to replace the optimization process with a feed-forward neural network that is trained to minimize the same objective~\cite{johnson2016perceptual,ulyanov2016texture,li2016precomputed}.
These feed-forward style transfer approaches are about three orders of magnitude faster than the optimization-based alternative, opening the door to real-time applications. 
Wang \etal~\cite{wang2016multimodal} enhanced the granularity of feed-forward style transfer with a multi-resolution architecture. Ulyanov~\etal~\cite{ulyanov2017improved} proposed ways to improve the quality and diversity of the generated samples. However, the above feed-forward methods are limited in the sense that each network is tied to a fixed style. To address this problem, Dumoulin \etal~\cite{dumoulin2017learned} introduced a single network that is able to encode $32$ styles and their interpolations. Concurrent to our work, Li~\etal~\cite{li2017diversified} proposed a feed-forward architecture that can synthesize up to $300$ textures and transfer $16$ styles. Still, the two methods above cannot adapt to arbitrary styles that are not observed during training. 

Very recently, Chen and Schmidt~\cite{chen2017fast} introduced a feed-forward method that can transfer arbitrary styles thanks to a style swap layer.
Given feature activations of the content and style images, the style swap layer replaces the content features with the closest-matching style features in a patch-by-patch manner.
Nevertheless, their style swap layer creates a new computational bottleneck: more than $95\%$ of the computation is spent on the style swap for $512\times 512$ input images. Our approach also permits arbitrary style transfer, while being $1$-$2$ orders of magnitude faster than~\cite{chen2017fast}.

Another central problem in style transfer is which style loss function to use. The original framework of Gatys~\etal~\cite{gatys2016image} matches styles by matching the second-order statistics between feature activations, captured by the Gram matrix. Other effective loss functions have been proposed, such as MRF loss~\cite{li2016combining}, adversarial loss~\cite{li2016precomputed}, histogram loss~\cite{wilmot2017stable}, CORAL loss~\cite{peng2017synthetic}, MMD loss~\cite{li2017demystifying}, and distance between channel-wise mean and variance~\cite{li2017demystifying}. 
Note that all the above loss functions aim to match some feature statistics between the style image and the synthesized image.

\vpara{Deep generative image modeling.} There are several alternative frameworks for image generation, including variational auto-encoders~\cite{kingma2014auto}, auto-regressive models~\cite{oord2016pixel}, and generative adversarial networks~(GANs)~\cite{goodfellow2014generative}. Remarkably, GANs have achieved the most impressive visual quality. Various improvements to the GAN framework have been proposed, such as conditional generation~\cite{reed2016generative,pix2pix2017}, multi-stage processing~\cite{denton2015deep,huang2017sgan}, and better training objectives~\cite{salimans2016improved,arjovsky2017wasserstein}. GANs have also been applied  to style transfer~\cite{li2016precomputed} and cross-domain image generation~\cite{taigman2017unsupervised,bousmalis2016unsupervised,pix2pix2017,liu2016coupled,liu2017unsupervised,kim2017discogan}.



\section{Background}
\subsection{Batch Normalization}

The seminal work of 	Ioffe and Szegedy~\cite{ioffe2015batch} introduced a batch normalization~(BN) layer that significantly ease the training of feed-forward networks by normalizing feature statistics.
BN layers are originally designed to accelerate training of discriminative networks, but have also been found effective in generative image modeling~\cite{radford2016unsupervised}.
Given an input batch $x \in \mathbb{R}^{N\times C\times H\times W}$, BN normalizes the mean and standard deviation for each individual feature channel:
\begin{equation}
\textrm{BN}(x)= \gamma\left(\frac{x-\mu(x)}{\sigma(x)}\right)+\beta
\end{equation}
where $\gamma, \beta \in \mathbb{R}^{C}$ are affine parameters learned from data; $\mu(x), \sigma(x) \in \mathbb{R}^{C}$ are the mean and standard deviation, computed across batch size and spatial dimensions independently for each feature channel:
\begin{equation}
\mu_{c}(x) = \frac{1}{NHW}\sum_{n=1}^{N}\sum_{h=1}^{H}\sum_{w=1}^{W}x_{nchw}
\end{equation}
\begin{equation}
\sigma_{c}(x) = \sqrt{\frac{1}{NHW}\sum_{n=1}^{N}\sum_{h=1}^{H}\sum_{w=1}^{W}(x_{nchw} - \mu_{c}(x))^{2} + \epsilon}
\end{equation}
BN uses mini-batch statistics during training and replace them with popular statistics during inference, introducing discrepancy between training and inference. Batch renormalization~\cite{ioffe2017batch} was recently proposed to address this issue by gradually using popular statistics during training.
As another interesting application of BN, Li \etal~\cite{li2017revisiting} found that BN can alleviate domain shifts by recomputing popular statistics in the target domain. Recently, several alternative normalization schemes have been proposed to extend BN's effectiveness to recurrent architectures~\cite{liao2016streaming,ba2016layer,salimans2016weight,cooijmans2017recurrent,laurent2016batch,ren2017normalizing}.

\begin{figure*}[!tb]
	\centering
	\subfigure[Trained with original images.]{
		\includegraphics[width=0.32\linewidth]{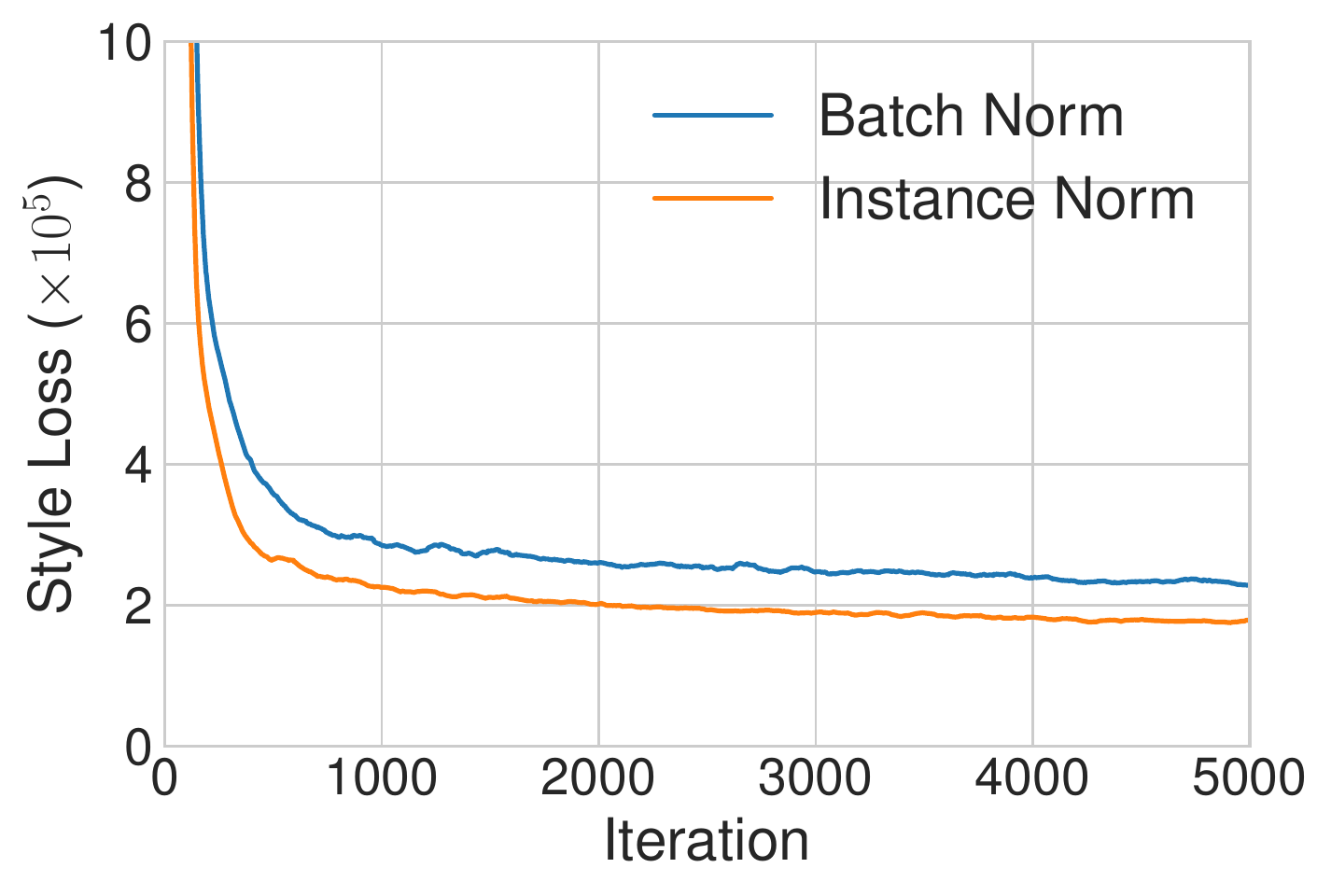}}
	\subfigure[Trained with contrast normalized images.]{
		\includegraphics[width=0.32\linewidth]{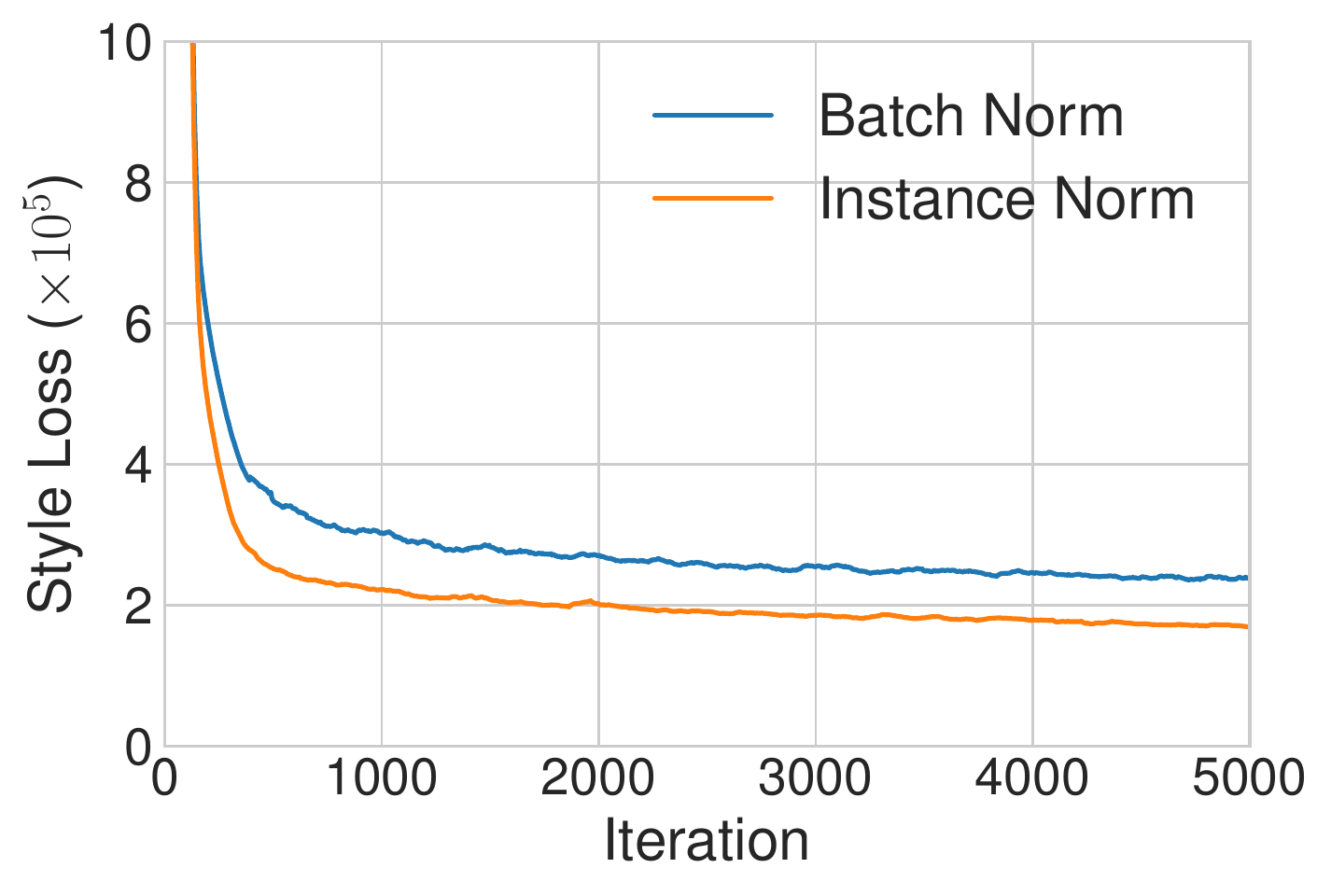}}
	\subfigure[Trained with style normalized images.]{
		\includegraphics[width=0.32\linewidth]{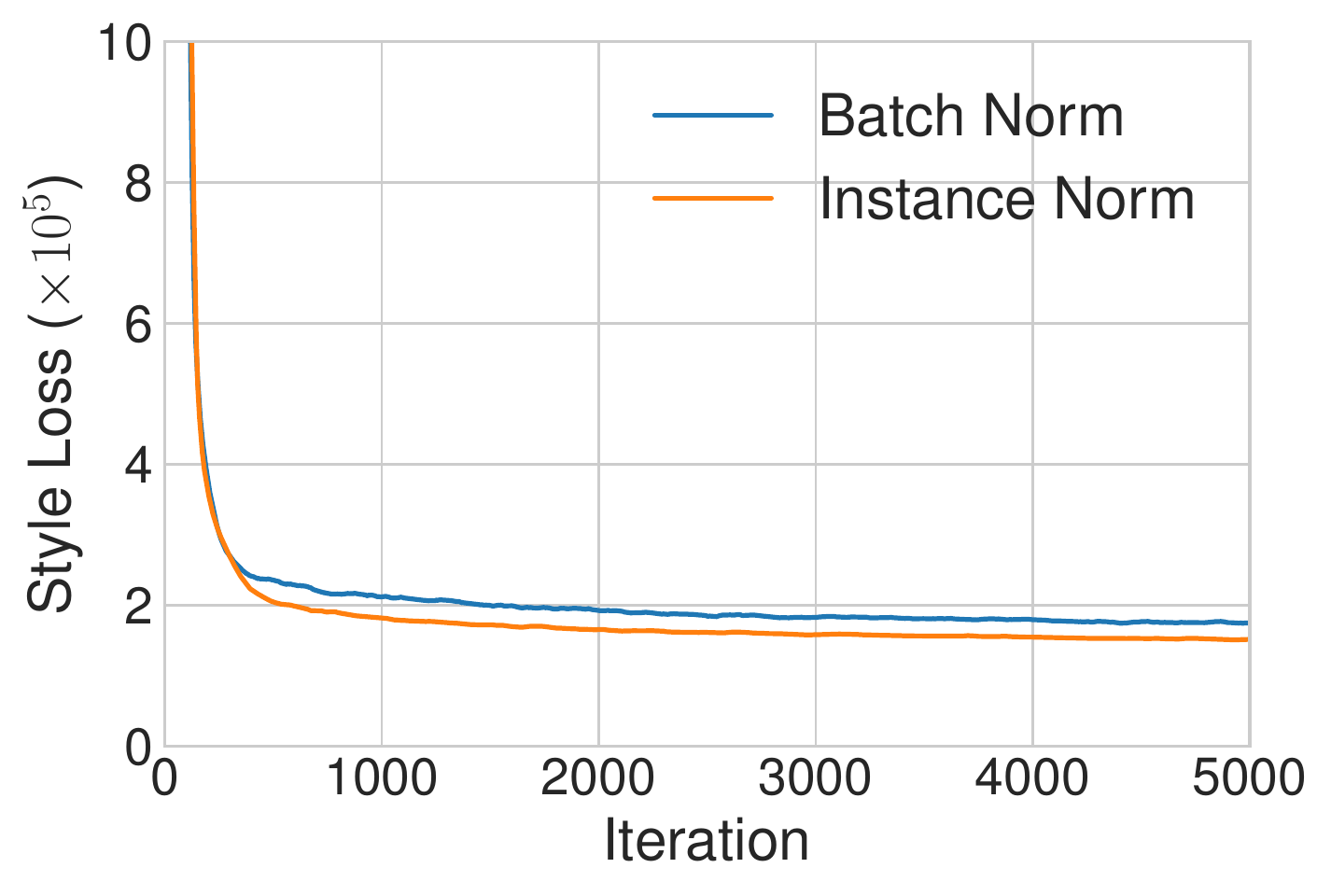}}
	\caption{To understand the reason for IN's effectiveness in style transfer, we train an IN model and a BN model with (a) original images in MS-COCO~\cite{lin2014microsoft}, (b) contrast normalized images, and (c) style normalized images using a pre-trained style transfer network~\cite{johnson2016perceptual}. The improvement brought by IN remains significant even when all training images are normalized to the same contrast, but are much smaller when all images are (approximately) normalized to the same style. Our results suggest that IN performs a kind of style normalization.} 
	\label{fig:in}
\end{figure*}

\subsection{Instance Normalization}

In the original feed-forward stylization method~\cite{ulyanov2016texture}, the style transfer network contains a BN layer after each convolutional layer. Surprisingly, Ulyanov \etal~\cite{ulyanov2017improved} found that significant improvement could be achieved simply by replacing BN layers with IN layers:
\begin{equation}
\textrm{IN}(x)= \gamma\left(\frac{x-\mu(x)}{\sigma(x)}\right)+\beta
\end{equation}

Different from BN layers, here $\mu(x)$ and $\sigma(x)$ are computed across spatial dimensions independently for each channel \emph{and each sample}:
\begin{equation}
\mu_{nc}(x) = \frac{1}{HW}\sum_{h=1}^{H}\sum_{w=1}^{W}x_{nchw}
\end{equation}
\begin{equation}
\sigma_{nc}(x) = \sqrt{\frac{1}{HW}\sum_{h=1}^{H}\sum_{w=1}^{W}(x_{nchw} - \mu_{nc}(x))^{2} + \epsilon}
\end{equation}

Another difference is that IN layers are applied at test time unchanged, whereas BN layers usually replace mini-batch statistics with population statistics. 

\subsection{Conditional Instance Normalization}
Instead of learning a single set of affine parameters $\gamma$ and $\beta$, Dumoulin \etal~\cite{dumoulin2017learned} proposed a \emph{conditional instance normalization}~(CIN) layer that learns a different set of parameters $\gamma^{s}$ and $\beta^{s}$ for each style $s$:
\begin{equation}
\textrm{CIN}(x; s)= \gamma^{s}\left(\frac{x-\mu(x)}{\sigma(x)}\right)+\beta^{s}
\end{equation}

During training, a style image together with its index $s$ are randomly chosen from a fixed set of styles $s\in \{1, 2, ..., S\}$~($S = 32$ in their experiments). 
The content image is then processed by a style transfer network in which the corresponding $\gamma^{s}$ and $\beta^{s}$ are used in the CIN layers. 
Surprisingly, the network can generate images in completely different styles by using the \emph{same} convolutional parameters but \emph{different}  affine parameters in IN layers.

Compared with a network without normalization layers, a network with CIN layers requires $2FS$ additional parameters, where $F$ is the total number of feature maps in the network~\cite{dumoulin2017learned}. Since the number of additional parameters scales linearly with the number of styles, it is challenging to extend their method to model a large number of styles~(\emph{e.g.}, tens of thousands). Also, their approach cannot adapt to arbitrary new styles without re-training the network.

\section{Interpreting Instance Normalization}
\label{sec:interpreting}

Despite the great success of (conditional) instance normalization, the reason why they work particularly well for style transfer remains elusive. Ulyanov~\etal~\cite{ulyanov2017improved} attribute the success of IN to its invariance to the contrast of the content image. However, IN takes place in the feature space, therefore it should have more profound impacts than a simple contrast normalization in the pixel space. Perhaps even more surprising is the fact that the affine parameters in IN can completely change the style of the output image.


It has been known that the convolutional feature statistics of a DNN can capture the style of an image~\cite{gatys2016image,li2016combining,li2017demystifying}. While Gatys~\etal~\cite{gatys2016image} use the second-order statistics as their optimization objective, Li \etal~\cite{li2017demystifying} recently showed that matching many other statistics, including channel-wise mean and variance, are also effective for style transfer. Motivated by these observations, we argue that instance normalization performs a form of \emph{style normalization} by normalizing feature statistics, namely the mean and variance. Although DNN serves as a image \emph{descriptor} in~\cite{gatys2016image,li2017demystifying}, we believe that the feature statistics of a \emph{generator} network can also control the style of the generated image.


We run the code of improved texture networks~\cite{ulyanov2017improved} to perform single-style transfer, with IN or BN layers. 
As expected, the model with IN converges faster than the BN model~(Fig.~\ref{fig:in}~(a)). To test the explanation in~\cite{ulyanov2017improved}, we then normalize all the training images to the same contrast by performing histogram equalization on the luminance channel. As shown in Fig.~\ref{fig:in}~(b), IN remains effective, suggesting the explanation in~\cite{ulyanov2017improved} to be incomplete. To verify our hypothesis, we normalize all the training images to the same style~(different from the target style) using a pre-trained style transfer network provided by~\cite{johnson2016perceptual}. According to Fig.~\ref{fig:in}~(c), the improvement brought by IN become much smaller when images are already style normalized. The remaining gap can explained by the fact that the style normalization with~\cite{johnson2016perceptual} is not perfect. Also, models with BN trained on style normalized images can converge as fast as models with IN trained on the original images. Our results indicate that IN does perform a kind of style normalization.

Since BN normalizes the feature statistics of a batch of samples instead of a single sample, it can be intuitively understood as normalizing a batch of samples to be centered around a single style. Each single sample, however, may still have different styles. This is undesirable when we want to transfer all images to the same style, 
as is the case in the original feed-forward style transfer algorithm~\cite{ulyanov2016texture}. 
Although the convolutional layers might learn to compensate the intra-batch style difference, it poses additional challenges for training.
On the other hand, IN can normalize the style of each individual sample to the target style. Training is facilitated because the rest of the network can focus on content manipulation while discarding the original style information. 
The reason behind the success of CIN also becomes clear: different affine parameters can normalize the feature statistics to different values, thereby normalizing the output image to different styles.

\section{Adaptive Instance Normalization}
If IN normalizes the input to a single style specified by the affine parameters, is it possible to adapt it to arbitrarily given styles by using adaptive affine transformations?
Here, we propose a simple extension to IN, which we call adaptive instance normalization~(AdaIN). 
AdaIN receives a content input $x$ and a style input $y$, and simply aligns the channel-wise mean and variance of $x$ to match those of $y$. Unlike BN, IN or CIN, AdaIN has no learnable affine parameters. Instead, it adaptively computes the affine parameters from the style input:
\begin{equation}
\textrm{AdaIN}(x, y)= \sigma(y)\left(\frac{x-\mu(x)}{\sigma(x)}\right)+\mu(y)
\end{equation}

in which we simply scale the normalized content input with $\sigma(y)$, and shift it with $\mu(y)$. Similar to IN, these statistics are computed across spatial locations. 

Intuitively, let us consider a feature channel that detects brushstrokes of a certain style. A style image with this kind of strokes will produce a high average activation for this feature. The output produced by AdaIN will have the same high average activation for this feature, while preserving the spatial structure of the content image. 
The brushstroke feature can be inverted to the image space with a feed-forward decoder, similar to~\cite{dosovitskiy2016inverting}.
The variance of this feature channel can encoder more subtle style information, which is also transferred to the AdaIN output and the final output image.

In short, AdaIN performs style transfer in the feature space by transferring feature statistics, specifically the channel-wise mean and variance. 
Our AdaIN layer plays a similar role as the style swap layer proposed in~\cite{chen2017fast}. While the style swap operation is very time-consuming and memory-consuming, our AdaIN layer is as simple as an IN layer, adding almost no computational cost.


\section{Experimental Setup}
\label{sec:setup}
Fig.~\ref{fig:arch} shows an overview of our style transfer network based on the proposed AdaIN layer.  Code and pre-trained models~(in Torch $7$~\cite{collobert2011torch7}) 
are available at: \url{https://github.com/xunhuang1995/AdaIN-style}
\begin{figure}[!t]
	\centering
	\includegraphics[width=1\linewidth]{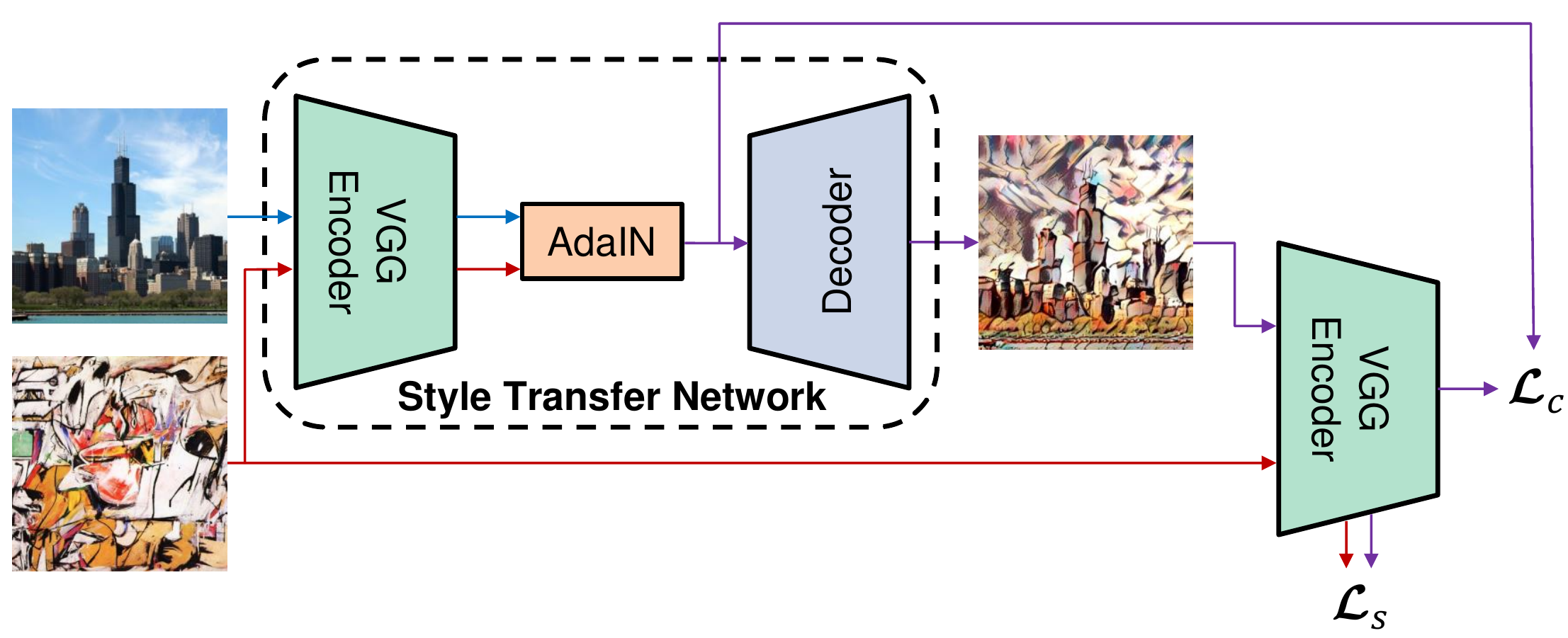}
	\caption{An overview of our style transfer algorithm. We use the first few layers of a fixed VGG-19 network to encode the content and style images. An AdaIN layer is used to perform style transfer in the feature space. A decoder is learned to invert the AdaIN output to the image spaces. We use the same VGG encoder to compute a content loss $\mathcal{L}_{c}$~(Equ.~\ref{equ:content}) and a style loss $\mathcal{L}_{s}$~(Equ.~\ref{equ:style}).}
	\label{fig:arch}
	\vspace{-0.2cm}
\end{figure}
\subsection{Architecture}
\label{subsec:arch}
Our style transfer network $T$ takes a content image $c$ and an arbitrary style image $s$ as inputs, and synthesizes an output image that recombines the content of the former and the style latter. We adopt a simple encoder-decoder architecture, in which the encoder $f$ is fixed to the first few layers~(up to $\texttt{relu4\_1}$) of a pre-trained VGG-19~\cite{simonyan2015very}. After encoding the content and style images in feature space, we feed both feature maps to an AdaIN layer that aligns the mean and variance of the content feature maps to those of the style feature maps, producing the target feature maps $t$: 
\begin{equation}
t = \textrm{AdaIN}(f(c), f(s))
\end{equation}

A randomly initialized decoder $g$ is trained to map $t$ back to the image space, generating the stylized image $T(c,s)$:
\begin{equation}
T(c, s) = g(t)
\end{equation}

The decoder mostly mirrors the encoder, with all pooling layers replaced by nearest up-sampling to reduce checkerboard effects. 
We use reflection padding in both $f$ and $g$ to avoid border artifacts. Another important architectural choice is whether the decoder should use instance, batch, or no normalization layers. As discussed in Sec.~\ref{sec:interpreting}, IN normalizes each sample to a single style while BN normalizes a batch of samples to be centered around a single style. Both are undesirable when we want the decoder to generate images in vastly different styles. Thus, we do \emph{not} use normalization layers in the decoder. In Sec.~\ref{sec:arbitrary} we will show that IN/BN layers in the decoder indeed hurt performance.



\subsection{Training}
\label{sec:training}

We train our network using MS-COCO~\cite{lin2014microsoft} as content images and a dataset of paintings mostly collected from WikiArt~\cite{wikiart2016} as style images, following the setting of~\cite{chen2017fast}. Each dataset contains roughly $80,000$ training examples. We use the adam optimizer~\cite{kingma2015adam} and a batch size of $8$ content-style image pairs. During training, we first resize the smallest dimension of both images to $512$ while preserving the aspect ratio, then randomly crop regions of size $256\times 256$. Since our network is fully convolutional, it can be applied to images of any size during testing.


Similar to~\cite{ulyanov2016texture,dumoulin2017learned,ulyanov2017improved}, we use the pre-trained VGG-19~\cite{simonyan2015very} to compute the loss function to train the decoder:
\begin{equation}
\label{equ:loss}
\mathcal{L} = \mathcal{L}_{c} + \lambda \mathcal{L}_{s}
\end{equation}

which is a weighted combination of the content loss $\mathcal{L}_{c}$ and the style loss $\mathcal{L}_{s}$ with the style loss weight $\lambda$. 
The content loss is the Euclidean distance between the target features and the features of the output image. We use the AdaIN output $t$ as the content target, instead of the commonly used feature responses of the content image. We find this leads to slightly faster convergence and also aligns with our goal of inverting the AdaIN output $t$.
\begin{equation}
\label{equ:content}
\mathcal{L}_{c} =\lVert f(g(t)) - t \rVert_{2}
\end{equation}

Since our AdaIN layer only transfers the mean and standard deviation of the style features, our style loss only matches these statistics. Although we find the commonly used Gram matrix loss can produce similar results, we match the IN statistics because it is conceptually cleaner. This style loss has also been explored by Li~\etal~\cite{li2017demystifying}.
\vspace{-0.1cm} 
\begin{multline}
\label{equ:style}
\mathcal{L}_{s} =\sum_{i=1}^{L}\lVert \mu(\phi_{i}(g(t))) - \mu(\phi_{i}(s)) \rVert_{2} \quad +\\
\sum_{i=1}^{L}\lVert \sigma(\phi_{i}(g(t))) - \sigma(\phi_{i}(s)) \rVert_{2}
\end{multline}

where each $\phi_{i}$ denotes a layer in VGG-19 used to compute the style loss. In our experiments we use $\texttt{relu1\_1}$, $\texttt{relu2\_1}$, $\texttt{relu3\_1}$, $\texttt{relu4\_1}$ layers with equal weights.

\section{Results}
\begin{figure}[!t]
\tiny
	\centering
	\subfigure[Style Loss]{
		\includegraphics[width=0.46\linewidth]{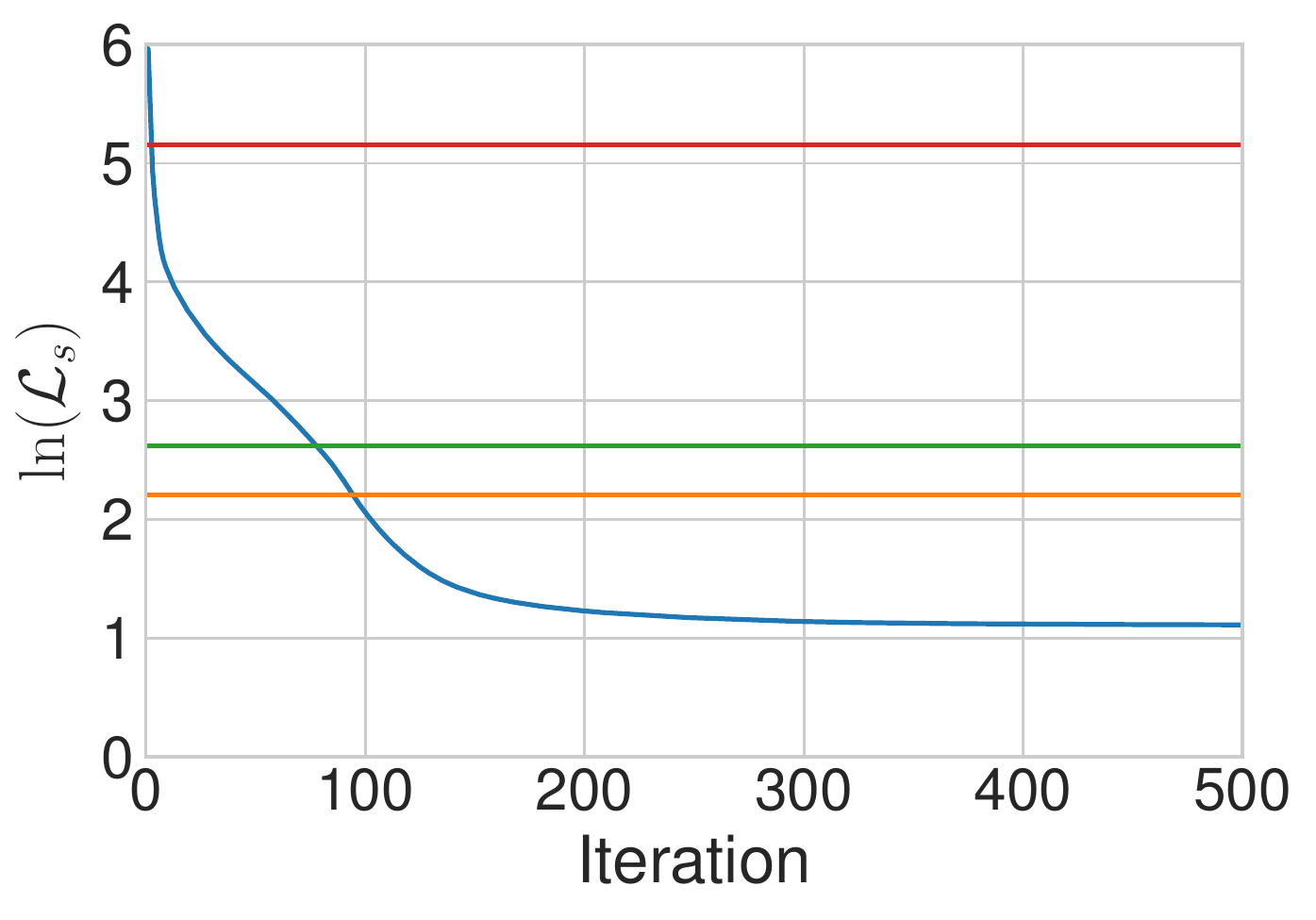}}
	\subfigure[Content Loss]{
		\includegraphics[width=0.482\linewidth]{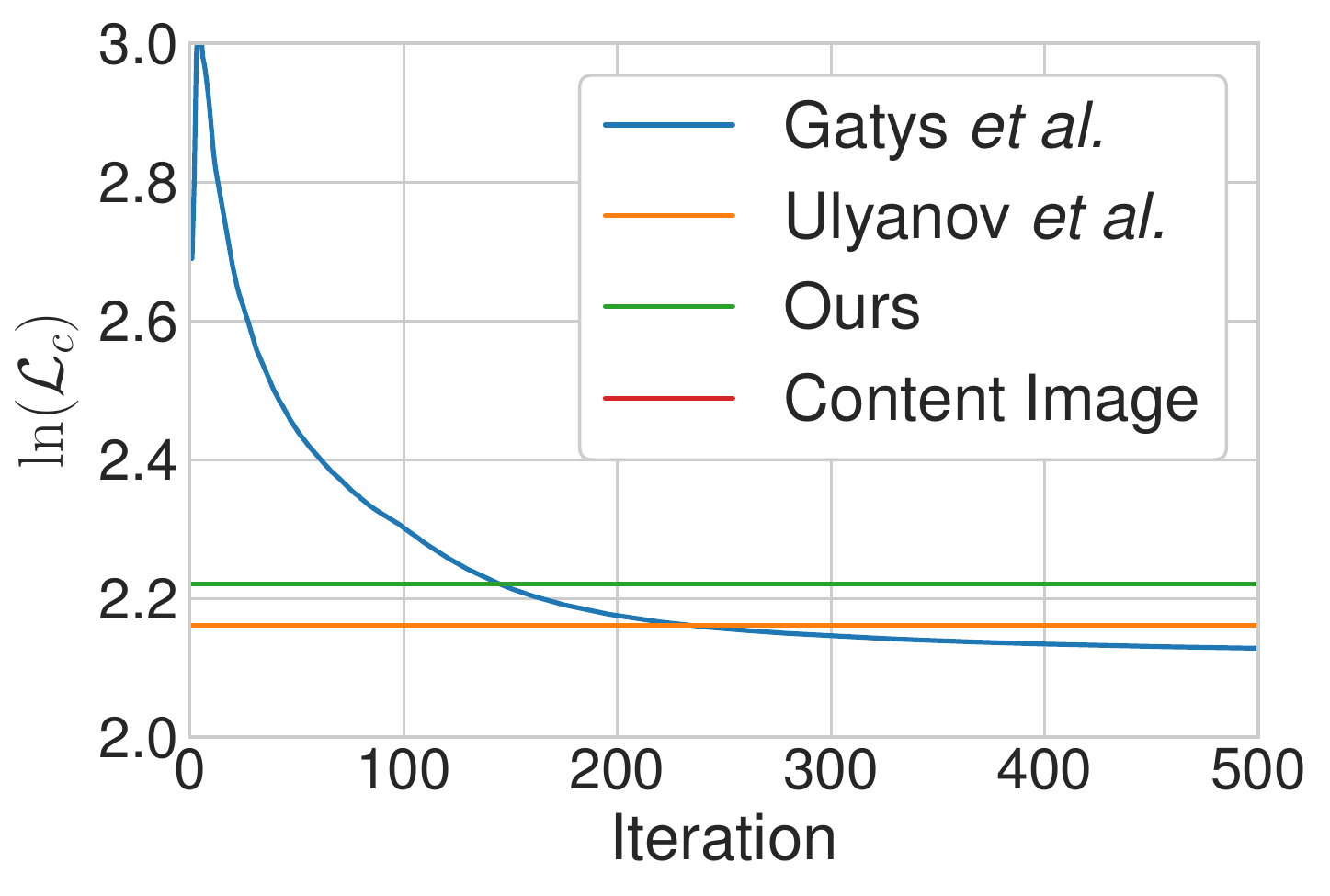}}

	\caption{Quantitative comparison of different methods in terms of style and content loss. Numbers are averaged over $10$ style images and $50$ content images randomly chosen from our test set.} 
	\label{fig:quantitative}
	\vspace{-0.3cm}
\end{figure}

\subsection{Comparison with other methods}
\label{sec:arbitrary}

In this subsection, we compare our approach with three types of style transfer methods: 1) the flexible but slow optimization-based method~\cite{gatys2016image}, 2) the fast feed-forward method restricted to a single style~\cite{ulyanov2017improved}, and 3) the flexible patch-based method of medium speed~\cite{chen2017fast}. 
If not mentioned otherwise, the results of compared methods are obtained by running their code with the default configurations.~\footnote{We run $500$ iterations of~\cite{gatys2016image} using Johnson's public implementation: \url{https://github.com/jcjohnson/neural-style}} For \cite{chen2017fast}, we use a pre-trained inverse network provided by the authors. All the test images are of size $512 \times 512$.

\begin{figure*}[!t]
\centering
\normalsize

\includegraphics[width=0.162\linewidth]{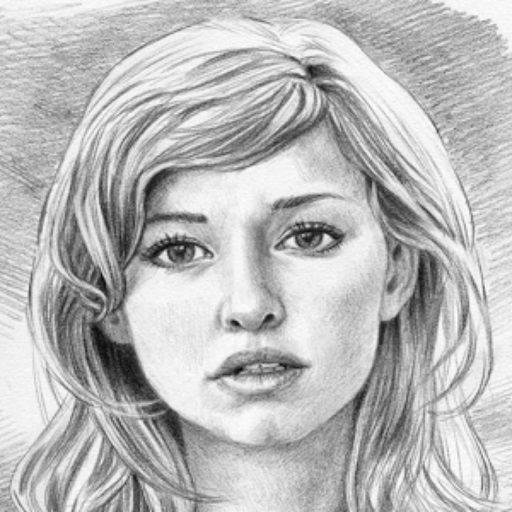}
\includegraphics[width=0.162\linewidth]{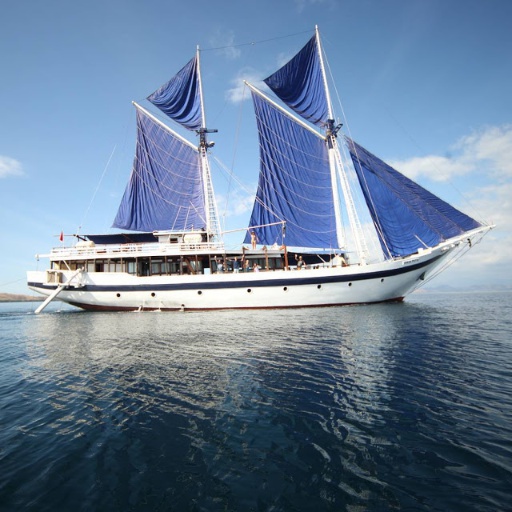}
\includegraphics[width=0.162\linewidth]{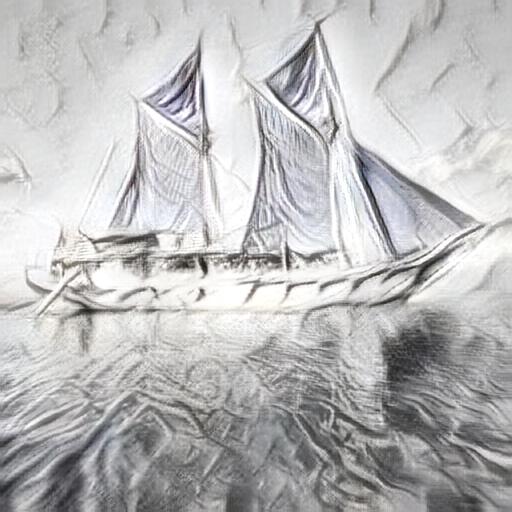}
\includegraphics[width=0.162\linewidth]{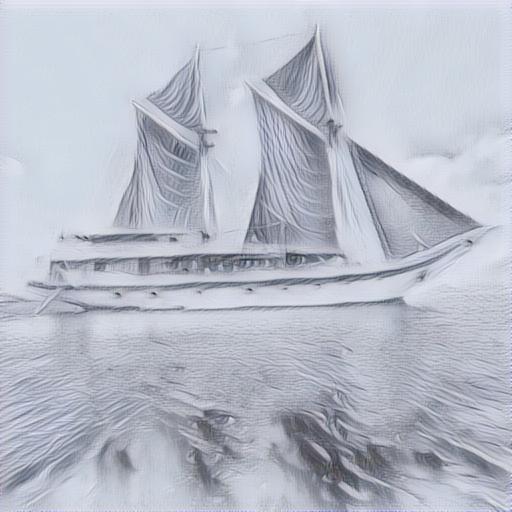}
\includegraphics[width=0.162\linewidth]{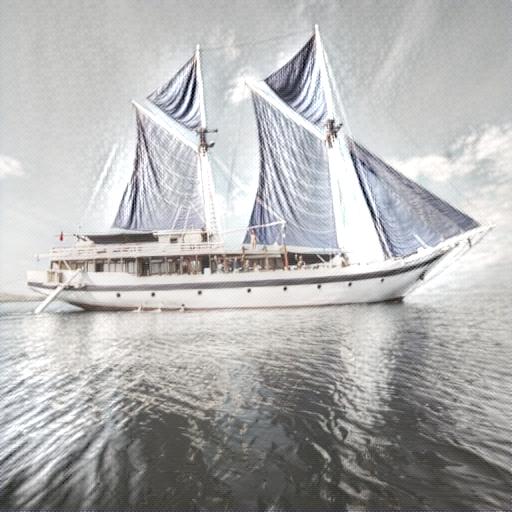}
\includegraphics[width=0.162\linewidth]{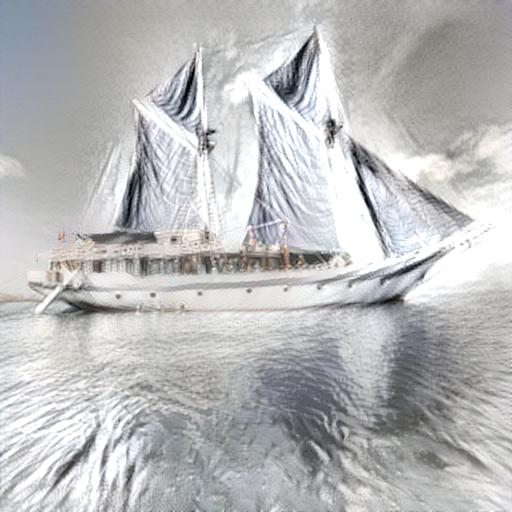}
\vspace{0.03cm}

\includegraphics[width=0.162\linewidth]{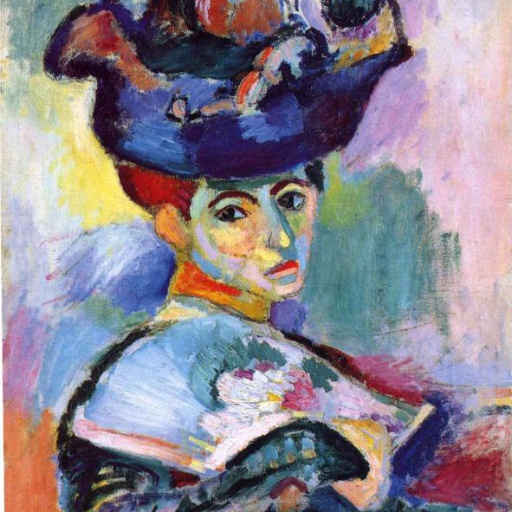}
\includegraphics[width=0.162\linewidth]{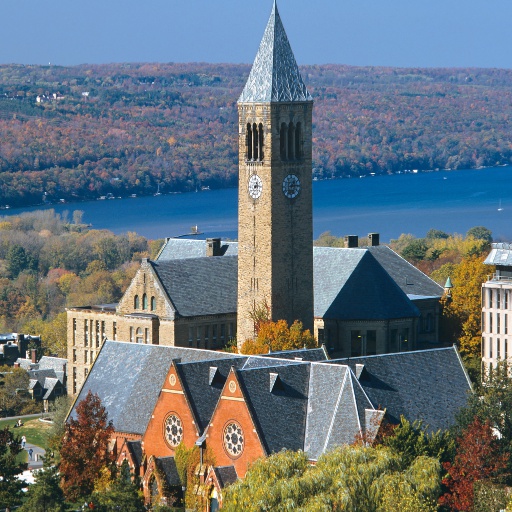}
\includegraphics[width=0.162\linewidth]{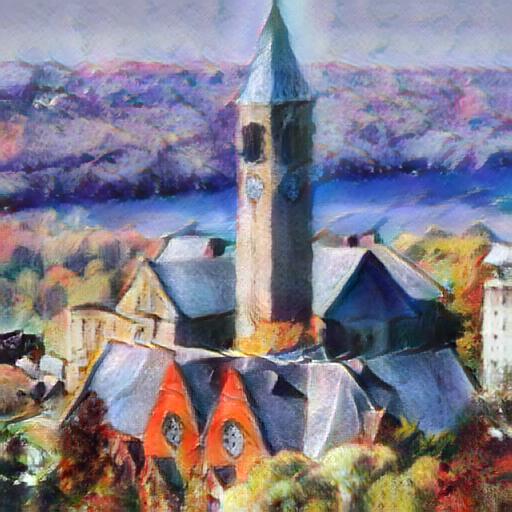}
\includegraphics[width=0.162\linewidth]{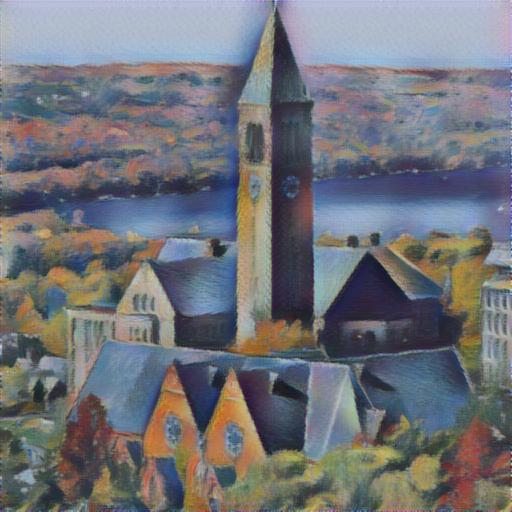}
\includegraphics[width=0.162\linewidth]{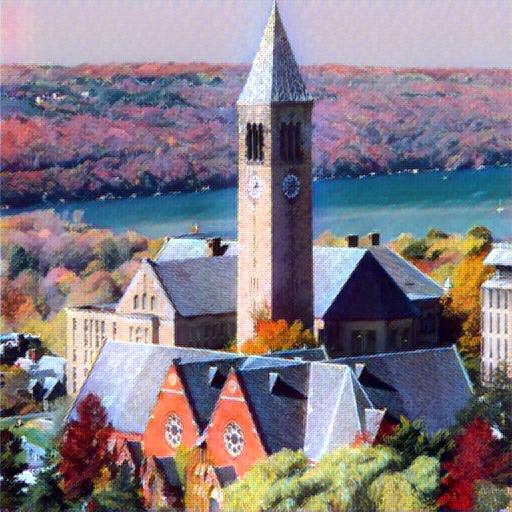}
\includegraphics[width=0.162\linewidth]{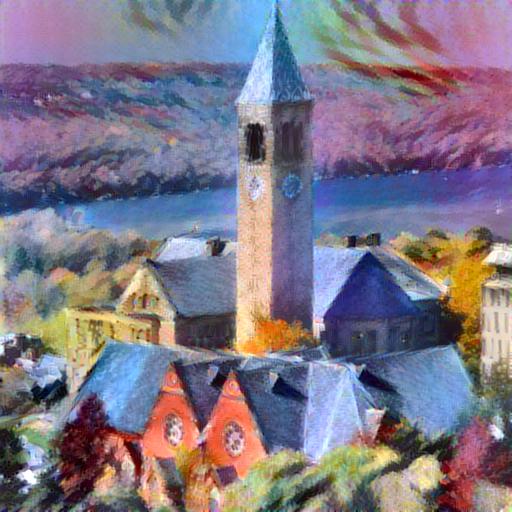}
\vspace{0.03cm}

\includegraphics[width=0.162\linewidth]{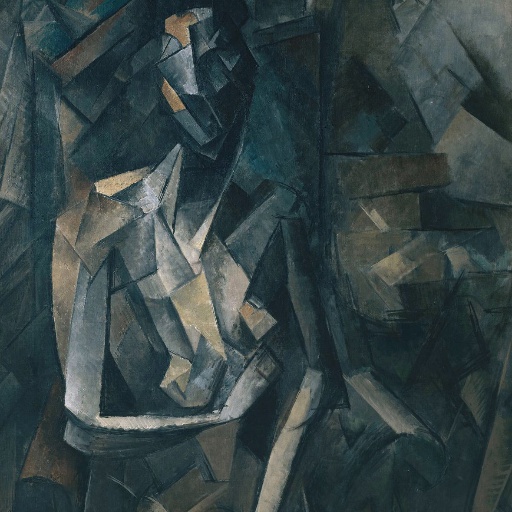}
\includegraphics[width=0.162\linewidth]{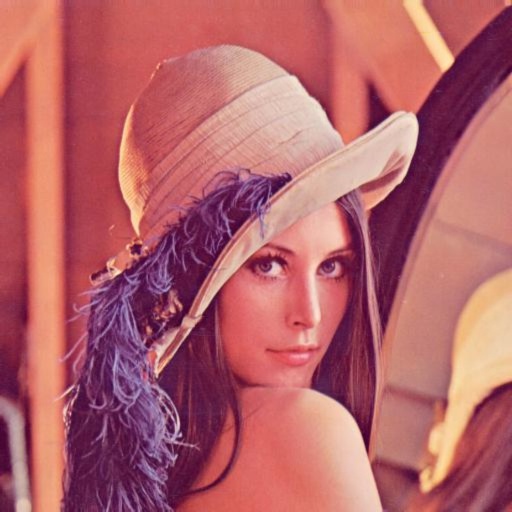}
\includegraphics[width=0.162\linewidth]{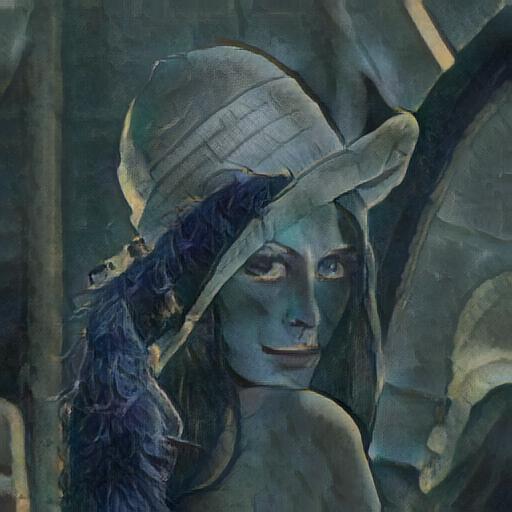}
\includegraphics[width=0.162\linewidth]{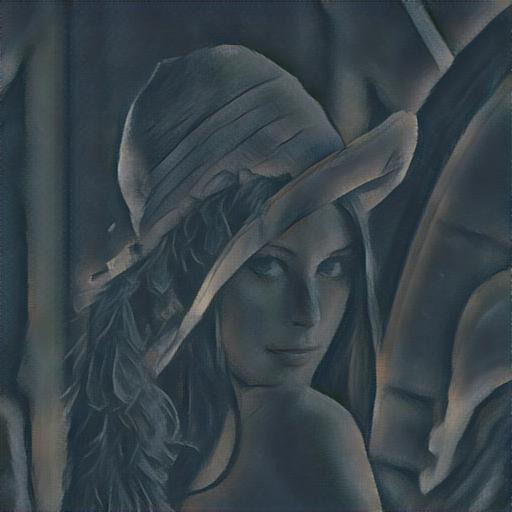}
\includegraphics[width=0.162\linewidth]{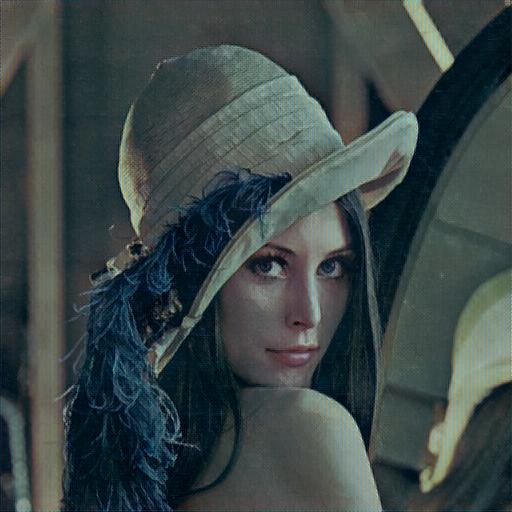}
\includegraphics[width=0.162\linewidth]{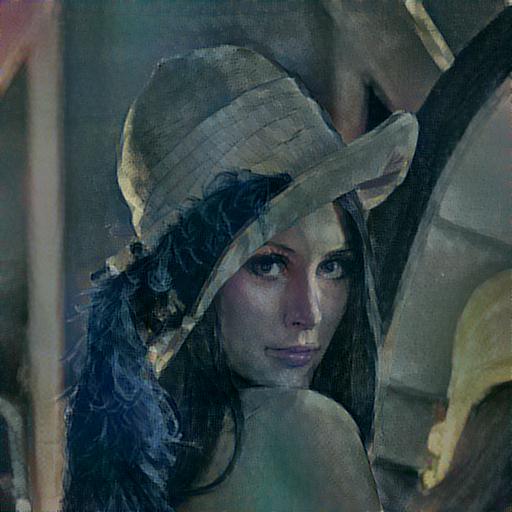}
\vspace{0.03cm}

%
\includegraphics[width=0.162\linewidth]{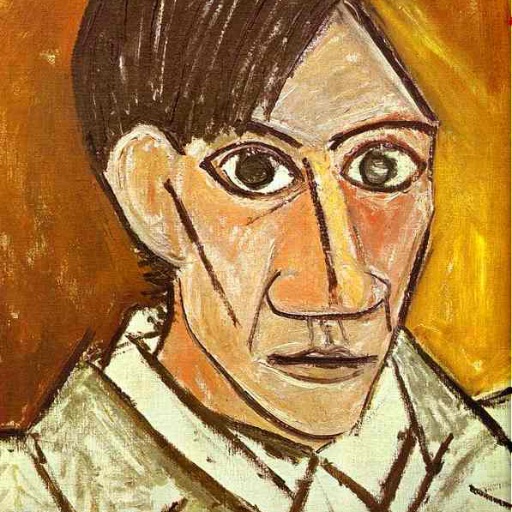}
\includegraphics[width=0.162\linewidth]{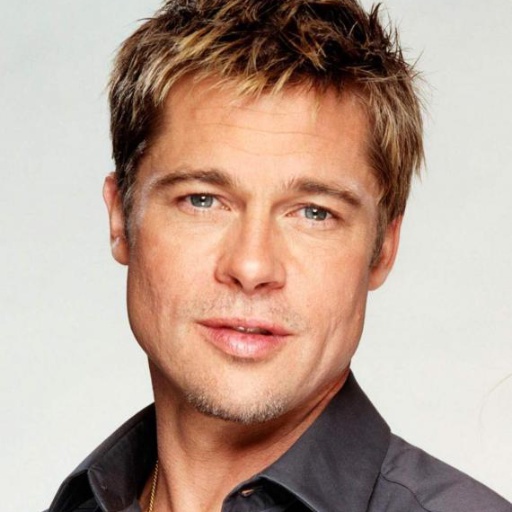}
\includegraphics[width=0.162\linewidth]{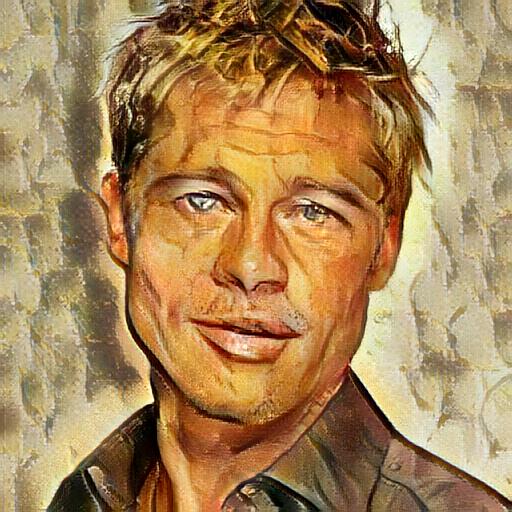}
\includegraphics[width=0.162\linewidth]{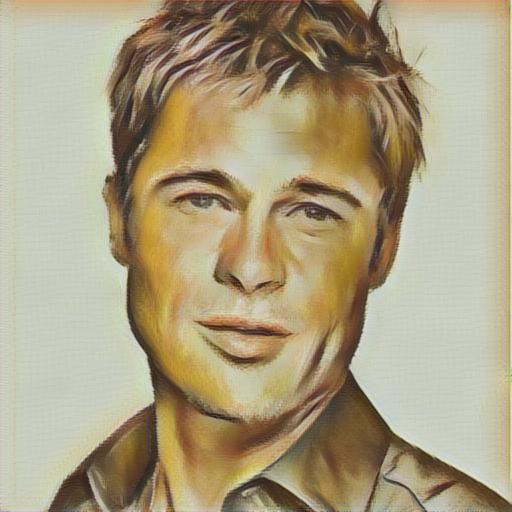}
\includegraphics[width=0.162\linewidth]{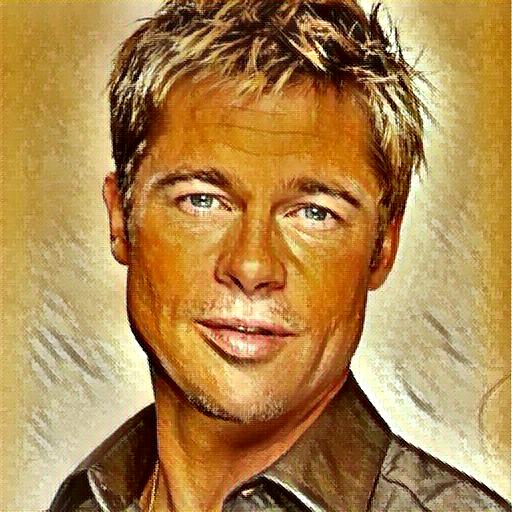}
\includegraphics[width=0.162\linewidth]{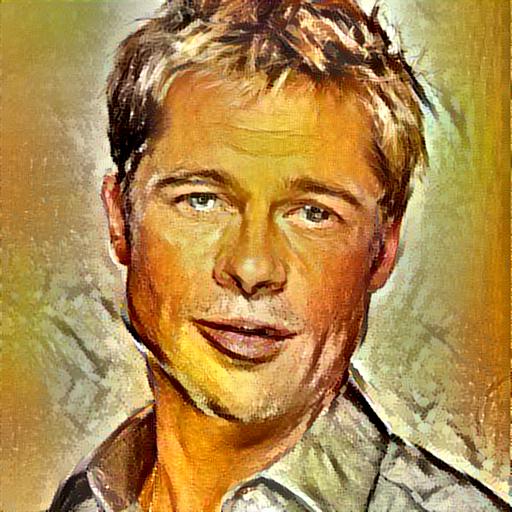}
\vspace{0.03cm}

\stackunder[5pt]{\includegraphics[width=0.162\linewidth]{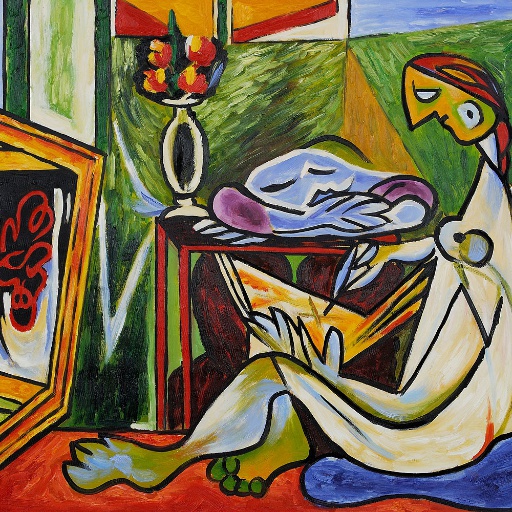}}{Style}
\stackunder[5pt]{\includegraphics[width=0.162\linewidth]{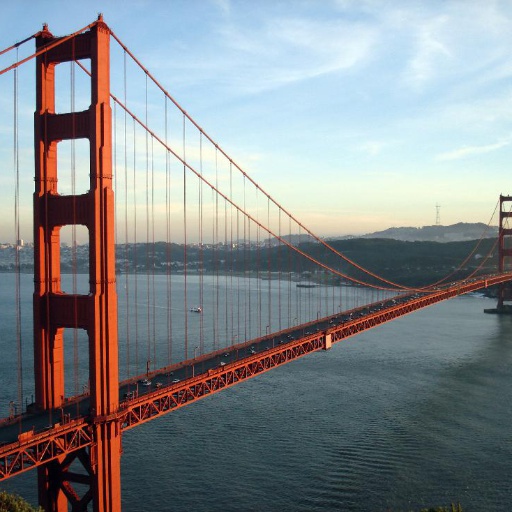}}{Content}
\stackunder[5pt]{\includegraphics[width=0.162\linewidth]{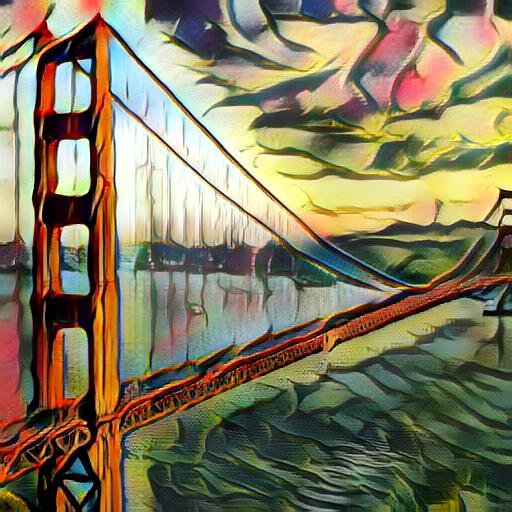}}{Ours}
\stackunder[5pt]{\includegraphics[width=0.162\linewidth]{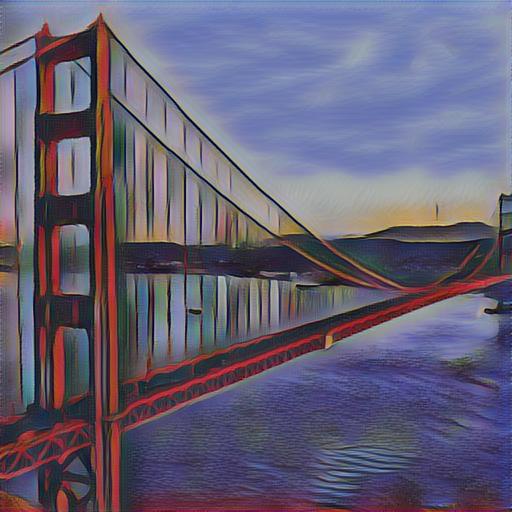}}{Chen and Schmidt}
\stackunder[5pt]{\includegraphics[width=0.162\linewidth]{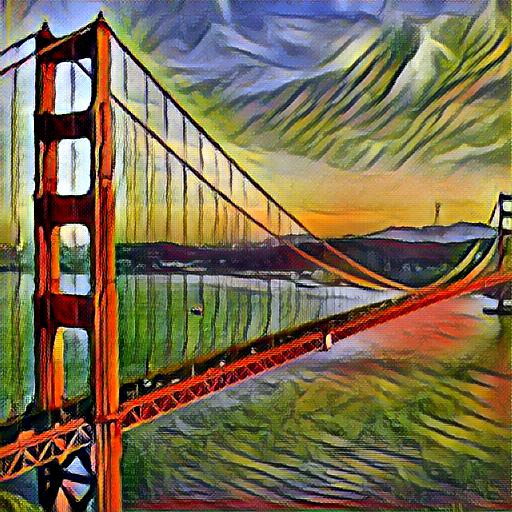}}{Ulyanov \etal}
\stackunder[5pt]{\includegraphics[width=0.162\linewidth]{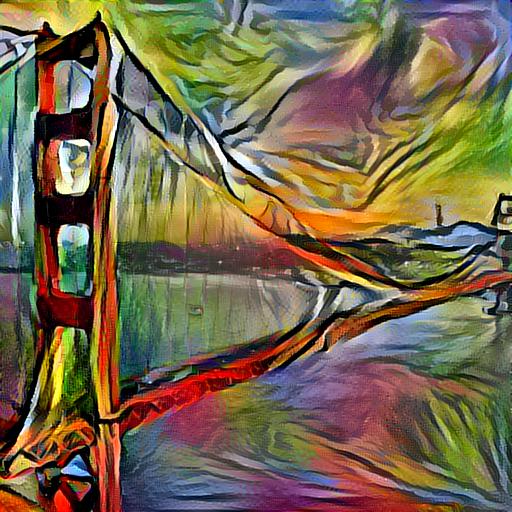}}{Gatys \etal}
\vspace{0.3cm}


\caption{Example style transfer results. All the tested content and style images are never observed by our network
during training.}
\label{fig:examples}
\end{figure*}

\vpara{Qualitative Examples.}  
In Fig.~\ref{fig:examples} we show example style transfer results generated by compared methods. 
Note that all the test style images are \emph{never} observed during the training of our model, while the results of~\cite{ulyanov2017improved} are obtained by fitting one network to each test style. 
Even so, the quality of our stylized images is quite competitive with \cite{ulyanov2017improved} and \cite{gatys2016image} for many images~(\emph{e.g.}, row $1,2,3$).
In some other cases~(\emph{e.g.}, row $5$) our method is slightly behind the quality of~\cite{ulyanov2017improved} and \cite{gatys2016image}. This is not unexpected, as we believe there is a three-way trade-off between speed, flexibility, and quality. 
Compared with~\cite{chen2017fast}, our method appears to transfer the style more faithfully for most compared images.
The last example clearly illustrates a major limitation of \cite{chen2017fast},
which attempts to match each content patch with the closest-matching style patch. However, if most content patches are matched to a few style patches that are not representative of the target style, the style transfer would fail. We thus argue that matching global feature statistics is a more general solution, 
although in some cases~(\emph{e.g.}, row $3$) the method of~\cite{chen2017fast} can also produce appealing results.


\vpara{Quantitative evaluations.}  
Does our algorithm trade off some quality for higher speed and flexibility, and if so by how much? 
To answer this question quantitatively, we compare our approach with the optimization-based method~\cite{gatys2016image} and the fast single-style transfer method~\cite{ulyanov2017improved} in terms of the content and style loss. Because our method uses a style loss based on IN statistics, we also modify the loss function in \cite{gatys2016image} and \cite{ulyanov2017improved} accordingly for a fair comparison~(their results in Fig.~\ref{fig:examples} are still obtained with the default Gram matrix loss). The content loss shown here is the same as in~\cite{ulyanov2017improved,gatys2016image}.
The numbers reported are averaged over $10$ style images and $50$ content images randomly chosen from the test set of the WikiArt dataset~\cite{wikiart2016} and MS-COCO~\cite{lin2014microsoft}. 


As shown in Fig.~\ref{fig:quantitative}, the average content and style loss of our synthesized images are slightly higher but comparable to the single-style transfer method of Ulyanov~\etal~\cite{ulyanov2017improved}. In particular, both our method and \cite{ulyanov2017improved} obtain a style loss similar to that of \cite{gatys2016image} between 50 and 100 iterations of optimization. 
This demonstrates the strong generalization ability of our approach, considering that our network has never seen the test styles during training while each network of~\cite{ulyanov2017improved} is specifically trained on a test style.
Also, note that our style loss is much smaller than that of the original content image.


\vspace{-0.1cm}
\vpara{Speed analysis.} 
Most of our computation is spent on content encoding, style encoding, and decoding, each roughly taking one third of the time.
In some application scenarios such as video processing, the style image needs to be encoded only once and AdaIN can use the stored style statistics to process all subsequent images. In some other cases~(\emph{e.g.}, transferring the same content to different styles), the computation spent on content encoding can be shared.

In Tab.~\ref{tab:speed} we compare the speed of our method with previous ones~\cite{gatys2016image,ulyanov2017improved,dumoulin2017learned,chen2017fast}.
Excluding the time for style encoding, our algorithm runs at $56$ and $15$ FPS for $256\times 256$ and $512\times 512$ images respectively, making it possible to process arbitrary user-uploaded styles in real-time. 
Among algorithms applicable to arbitrary styles, our method is nearly $3$ orders of magnitude faster than~\cite{gatys2016image} and $1$-$2$ orders of magnitude faster than~\cite{chen2017fast}. The speed improvement over~\cite{chen2017fast} is particularly significant for images of higher resolution, since the style swap layer in~\cite{chen2017fast} does not scale well to high resolution style images. 
Moreover, our approach achieves comparable speed to feed-forward methods limited to a few styles~\cite{ulyanov2017improved,dumoulin2017learned}. The slightly longer processing time of our method is mainly due to our larger VGG-based network, instead of methodological limitations. With a more efficient architecture, our speed can be further improved.

\begin{table}[!tb]
\addtolength{\tabcolsep}{-3pt}
\renewcommand\arraystretch{1.3}
\centering
\small
\begin{tabular}{l|l|l|c}
Method & Time~($256$px) & Time~($512$px)  & \# Styles \\ \hhline{=|=|=|=}
Gatys \etal      &   $14.17$~($14.19$)   &   $46.75$~($46.79$)  & $\textcolor{blue}{\infty}$ \\ 
Chen and Schmidt      &    $0.171$~($0.407$)  &  $3.214$~($4.144$)    & $\textcolor{blue}{\infty}$ \\ 
Ulyanov \etal      &   $\textbf{0.011}$~(N/A)  &   $\textbf{0.038}$~(N/A)  & 1  \\ 
Dumoulin \etal      &   $\textbf{0.011}$~(N/A)  &   $\textbf{0.038}$~(N/A)   & 32 \\ 
Ours      &   $\textbf{0.018}$~($0.027$)  &   $\textbf{0.065}$~($0.098$)   &    $\textcolor{blue}{\infty}$  \\ 
\end{tabular}\\
\vspace{0.2cm}
\caption{Speed comparison~(in seconds) for $256\times 256$ and $512\times 512$ images. Our approach achieves comparable speed to methods limited to a small number styles~\cite{ulyanov2017improved,dumoulin2017learned}, while being much faster than other existing algorithms applicable to arbitrary styles~\cite{gatys2016image,chen2017fast}. We show the processing time both excluding and including~(in parenthesis) the style encoding procedure. 
Results are obtained with a Pascal Titan X GPU and averaged over $100$ images.}
\vspace{-0.3cm}
\label{tab:speed}
\end{table}

\subsection{Additional experiments.} 
In this subsection, we conduct experiments to justify our important architectural choices. We denote our approach described in Sec.~\ref{sec:setup} as Enc-AdaIN-Dec. We experiment with a model named Enc-Concat-Dec that replaces AdaIN with concatenation, which is a natural baseline strategy to combine information from the content and style images. In addition, we run models with BN/IN layers in the decoder, denoted as Enc-AdaIN-BNDec and Enc-AdaIN-INDec respectively. Other training settings are kept the same.

\begin{figure}[!t]
	\centering
	\subfigure[Style]{
		\includegraphics[width=0.322\linewidth]{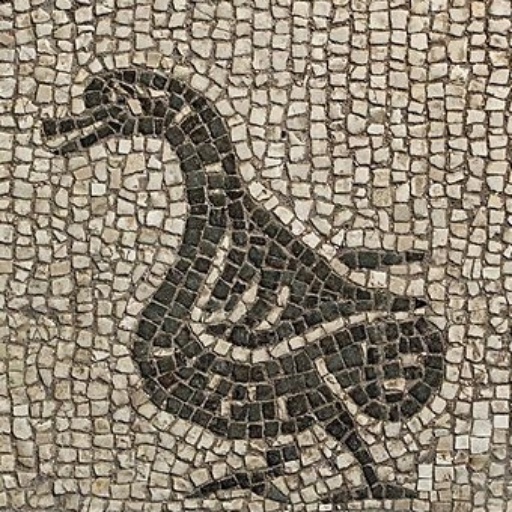}}\hfill
	\subfigure[Content]{
		\includegraphics[width=0.322\linewidth]{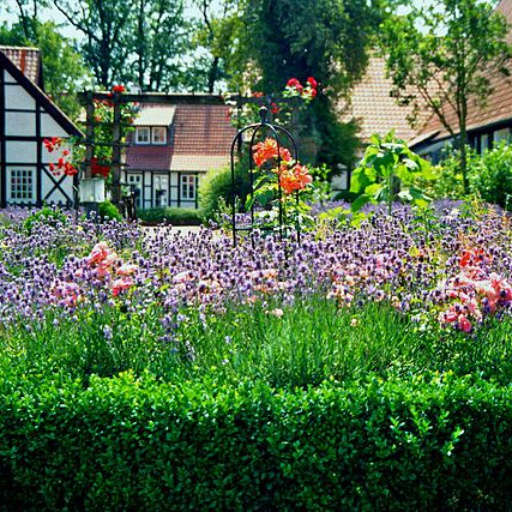}}\hfill
	\subfigure[Enc-AdaIN-Dec]{
		\includegraphics[width=0.322\linewidth]{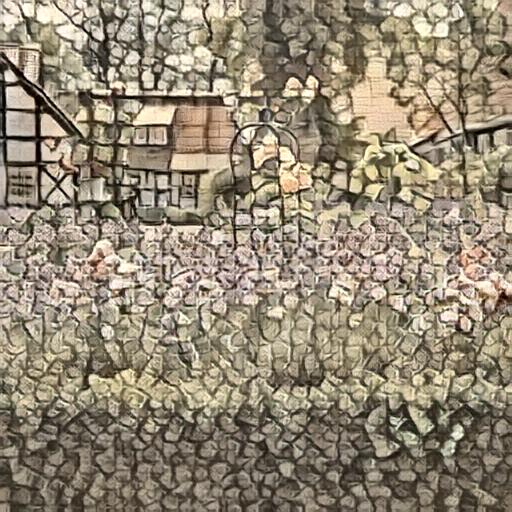}}
	\subfigure[Enc-Concat-Dec]{
		\includegraphics[width=0.322\linewidth]{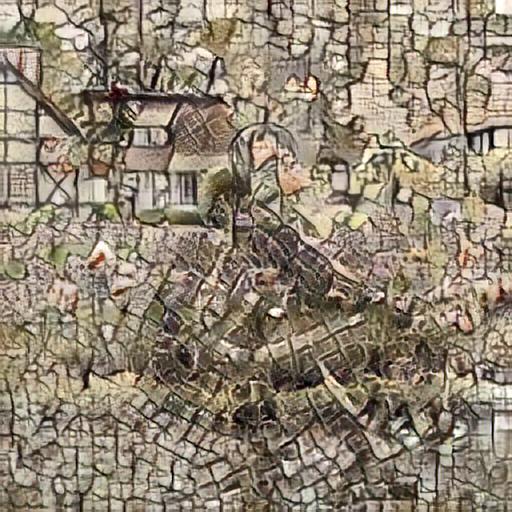}}\hfill	
	\subfigure[Enc-AdaIN-BNDec]{
		\includegraphics[width=0.322\linewidth]{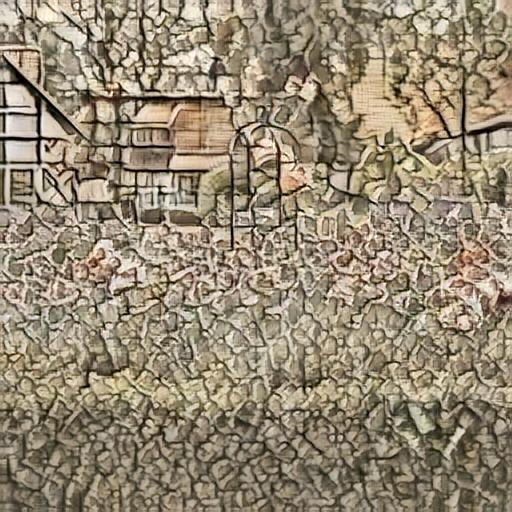}}\hfill
	\subfigure[Enc-AdaIN-INDec]{
		\includegraphics[width=0.322\linewidth]{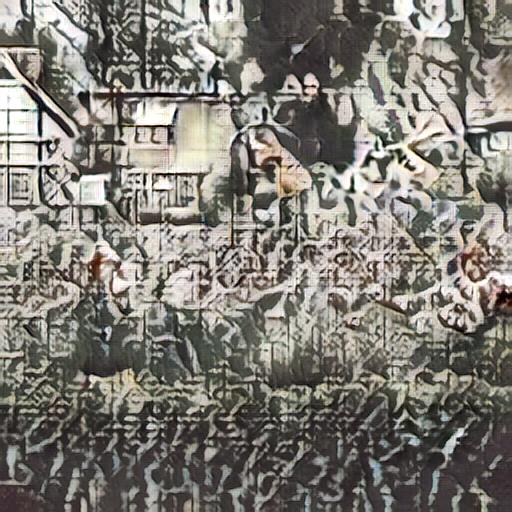}}\\
		\vspace{0.1cm}
	\caption{Comparison with baselines. AdaIN is much more effective than concatenation in fusing the content and style information. Also, it is important \emph{not} to use BN or IN layers in the decoder.} 
	\label{fig:baselines}
\end{figure}

\begin{figure}[!t]
	\centering
	\includegraphics[width=1\linewidth]{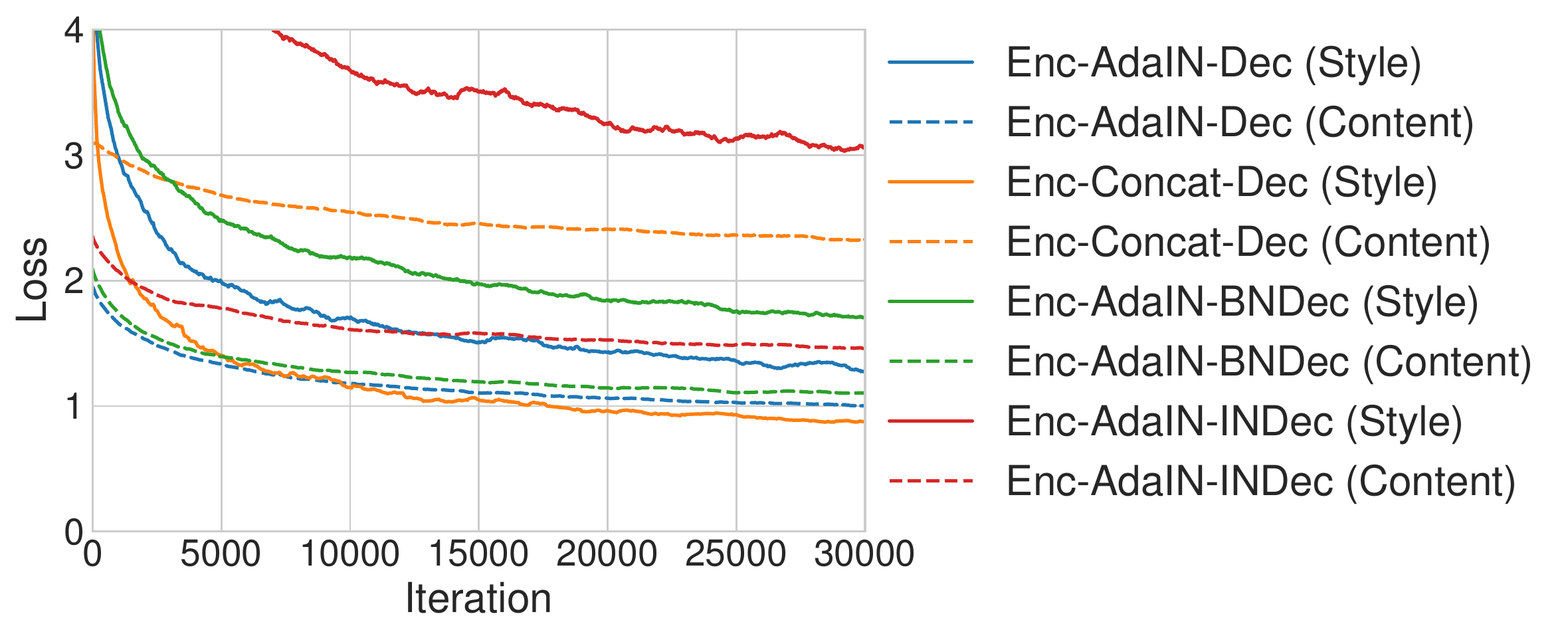}
	\caption{Training curves of style and content loss.}
	\label{fig:curves}
	\vspace{-0.4cm}
\end{figure}

In Fig.~\ref{fig:baselines} and \ref{fig:curves}, we show examples and training curves of the compared methods. In the image generated by the Enc-Concat-Dec baseline~(Fig.~\ref{fig:baselines}~(d)), the object contours of the style image can be clearly observed, suggesting that the network fails to disentangle the style information from the content of the style image. This is also consistent with  Fig.~\ref{fig:curves}, where Enc-Concat-Dec can reach low style loss but fail to decrease the content loss. Models with BN/IN layers also obtain qualitatively worse results and consistently higher losses. The results with IN layers are especially poor. This once again verifies our claim that IN layers tend to normalize the output to a single style and thus should be avoided when we want to generate images in different styles.






\subsection{Runtime controls}

To further highlight the flexibility of our method, we show that our style transfer network allows users to control the degree of stylization, interpolate between different styles, transfer styles while preserving colors, and use different styles in different spatial regions. Note that all these controls are only applied at runtime using the same network, without any modification to the training procedure. 

\vpara{Content-style trade-off.} The degree of style transfer can be controlled during training by adjusting the style weight $\lambda$ in Eqa.~\ref{equ:loss}. In addition, our method allows content-style trade-off at test time by interpolating between feature maps that are fed to the decoder. 
Note that this is equivalent to interpolating between the affine parameters of AdaIN.								\begin{equation}
\label{equ:tradeoff}
T(c, s, \alpha)=g((1-\alpha) f(c) + \alpha \textrm{AdaIN}(f(c), f(s)))
\end{equation}

\begin{figure*}[!htb]
\centering
\small

\stackunder[5pt]{\includegraphics[width=0.162\linewidth]{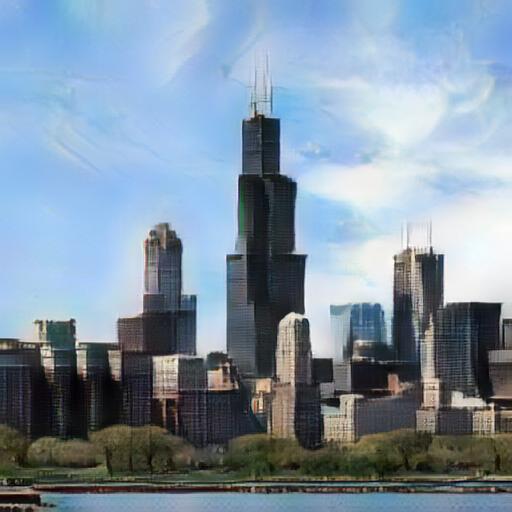}}{$\alpha = 0$}
\stackunder[5pt]{\includegraphics[width=0.162\linewidth]{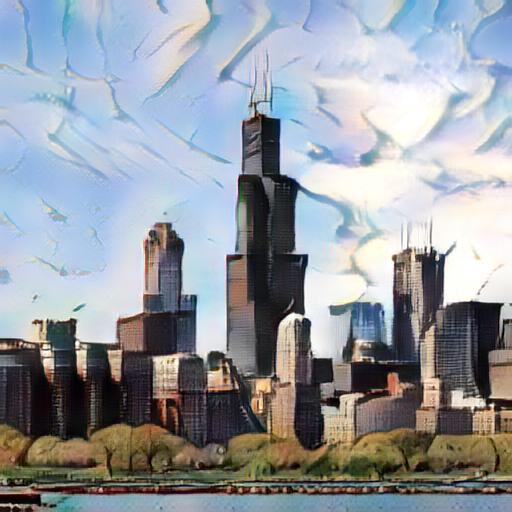}}{$\alpha = 0.25$}
\stackunder[5pt]{\includegraphics[width=0.162\linewidth]{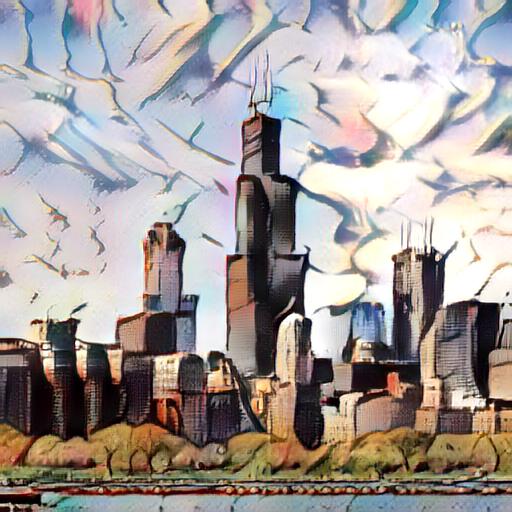}}{$\alpha = 0.5$}
\stackunder[5pt]{\includegraphics[width=0.162\linewidth]{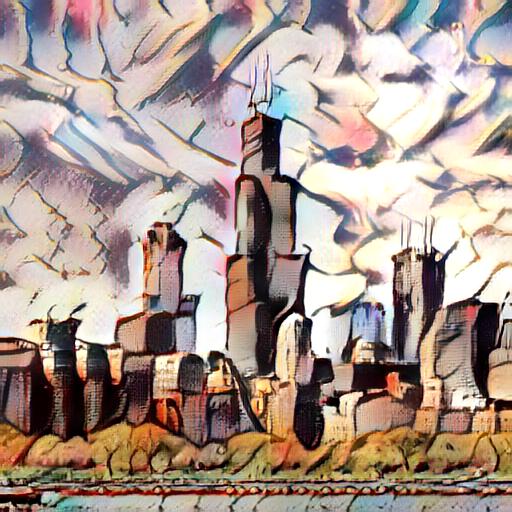}}{$\alpha = 0.75$}
\stackunder[5pt]{\includegraphics[width=0.162\linewidth]{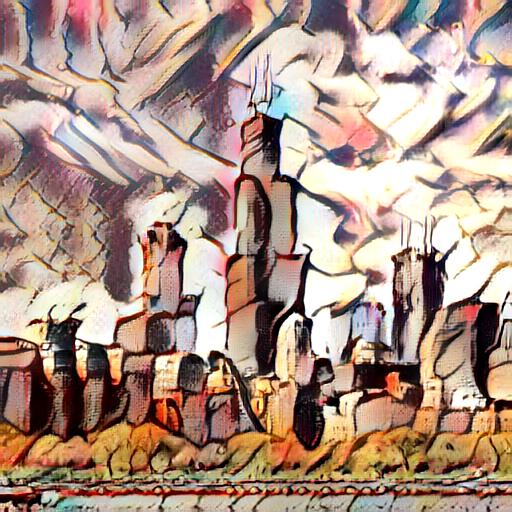}}{$\alpha = 1$}
\stackunder[5pt]{\includegraphics[width=0.162\linewidth]{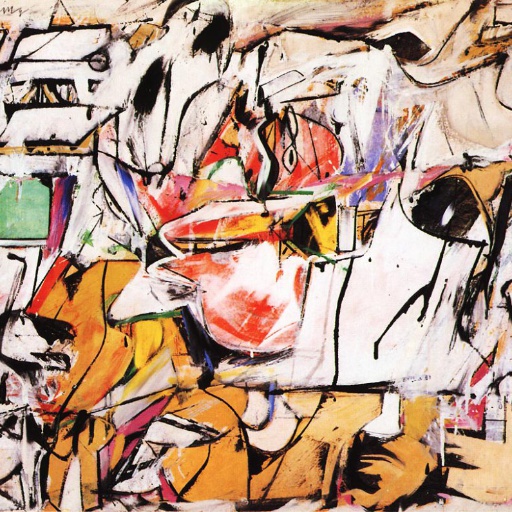}}{Style}\\
\vspace{0.2cm}

\caption{Content-style trade-off. At runtime, we can control the balance between content and style by changing the weight $\alpha$ in Equ.~\ref{equ:tradeoff}.}
\label{fig:tradeoff}
\vspace{-0.2cm}
\end{figure*}

The network tries to faithfully reconstruct the content image when $\alpha=0$, and to synthesize the most stylized image when $\alpha=1$. 
As shown in Fig.~\ref{fig:tradeoff}, a smooth transition between content-similarity and style-similarity can be observed by changing $\alpha$ from $0$ to $1$.

\begin{figure}[!tb]
\centering
\small
\includegraphics[width=1\linewidth]{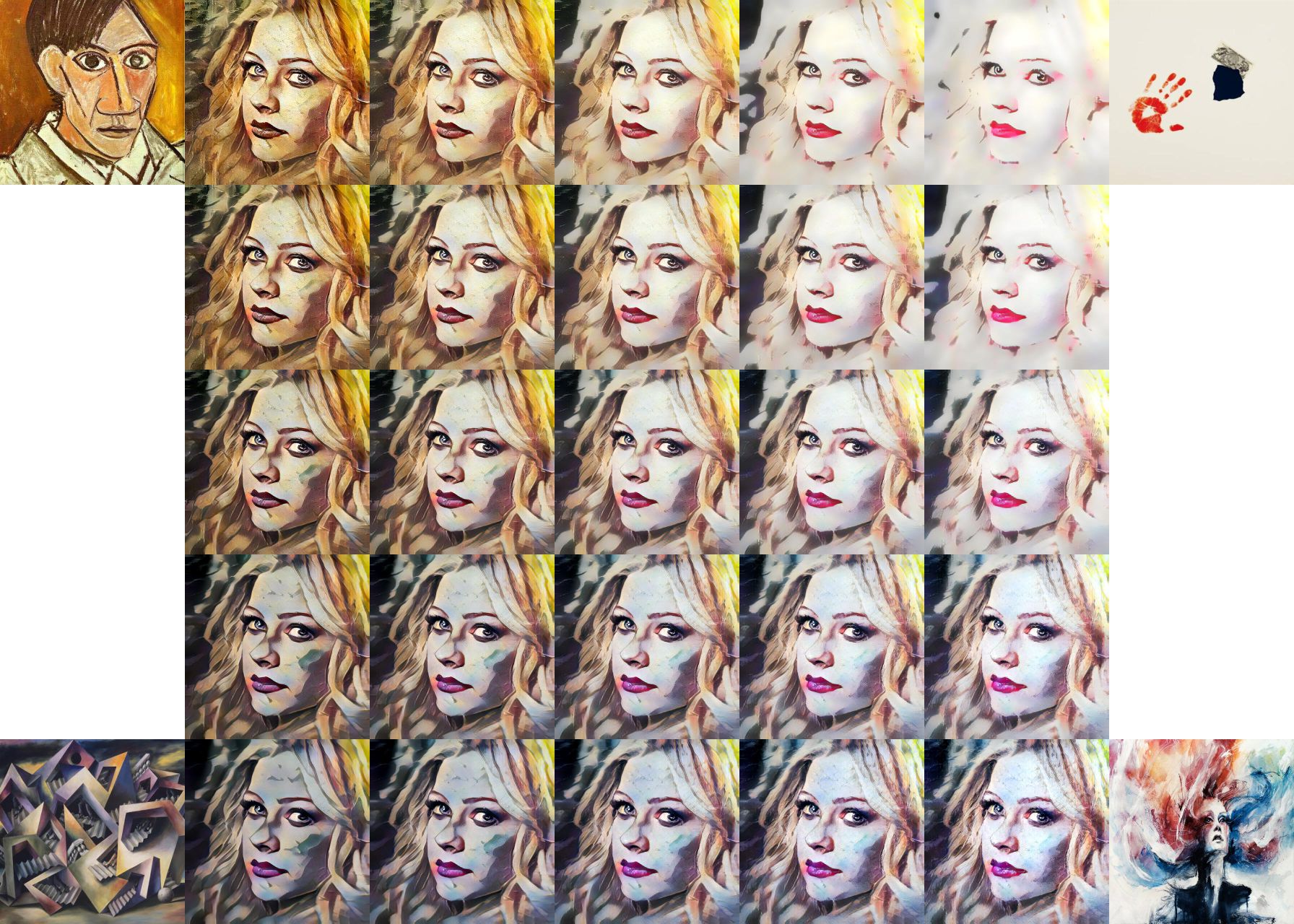}\\
\vspace{0.1cm}
\caption{Style interpolation. By feeding the decoder with a convex combination of feature maps transferred to different styles via AdaIN~(Equ.~\ref{equ:interp}), we can interpolate between arbitrary new styles. }
\vspace{-0.3cm}
\label{fig:interp}
\end{figure}

\vpara{Style interpolation.} To interpolate between a set of $K$ style images $s_{1}, s_{2}, ... ,s_{K}$ with corresponding weights $w_{1}, w_{2}, ... ,w_{K}$ such that $\sum_{k=1}^{K}w_{k} = 1$, we similarly interpolate between feature maps~(results shown in Fig.~\ref{fig:interp}):
\begin{equation}
\label{equ:interp}
T(c, s_{1, 2, ...K}, w_{1, 2, ...K})=g(\sum_{k=1}^{K}w_{k}\textrm{AdaIN}(f(c), f(s_{k})))
\end{equation}


%
%
%



\begin{figure}[!tb]
\centering
\small
%

%
\begin{overpic}[width=0.492\linewidth]%
	{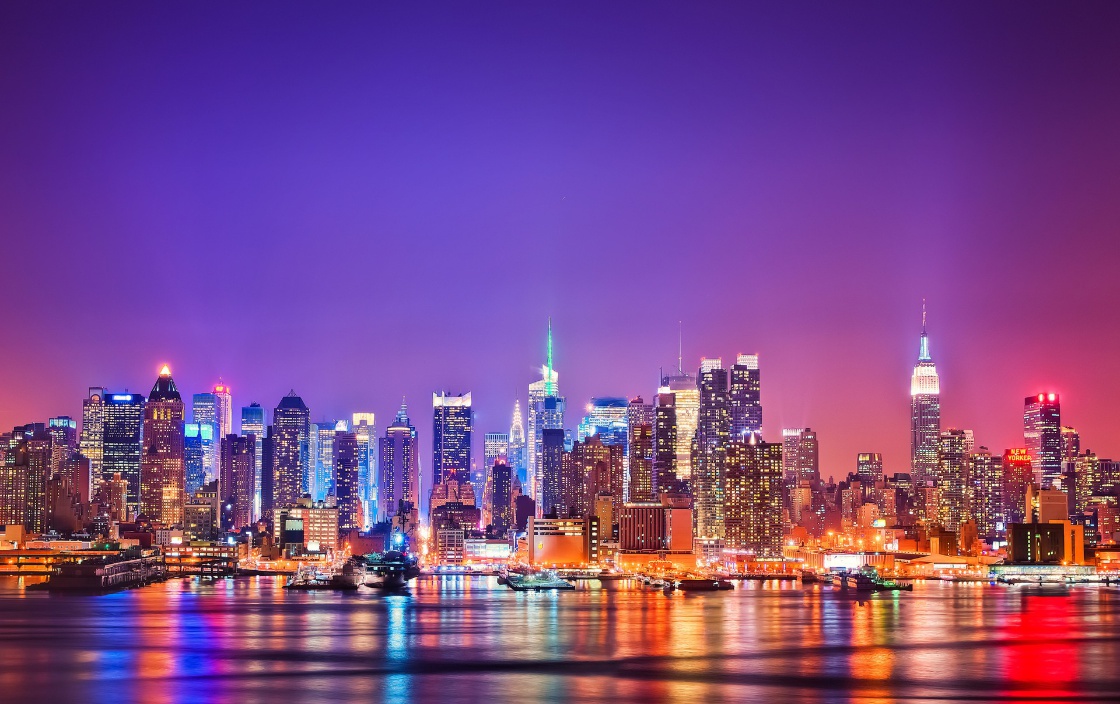}
\put(81.1,46.7){\includegraphics[width=0.15\linewidth]%
{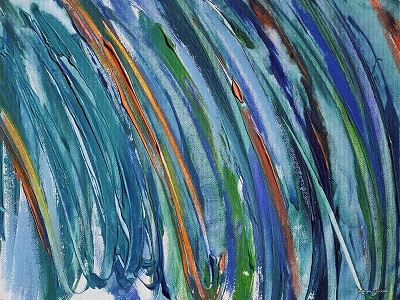}}
\end{overpic}
\hspace{0.03cm}
\begin{overpic}[width=0.492\linewidth]%
{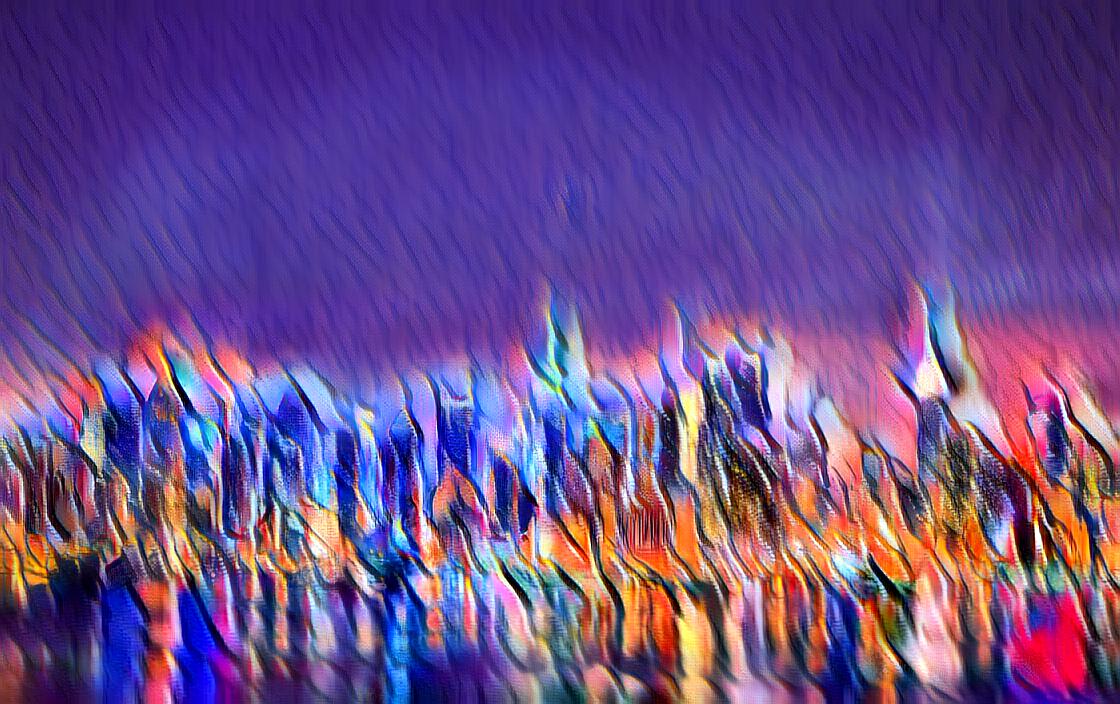}
\end{overpic}
\\
\vspace{0.05cm}

\caption{Color control. Left: content and style images. Right: color-preserved style transfer result.}
\label{fig:color}
\end{figure}

\vpara{Spatial and color control.} Gatys \etal~\cite{gatys2017controlling} recently introduced user controls over color information and spatial locations of style transfer, which can be easily incorporated into our framework. To preserve the color of the content image, we first match the color distribution of the style image to that of the content image~(similar to~\cite{gatys2017controlling}), then perform a normal style transfer using the color-aligned style image as the style input. Examples results are  shown in Fig.~\ref{fig:color}.

\begin{figure}[!tb]
\centering
\small
\noindent
\begin{minipage}{.32\linewidth}
\includegraphics[width=1\linewidth]{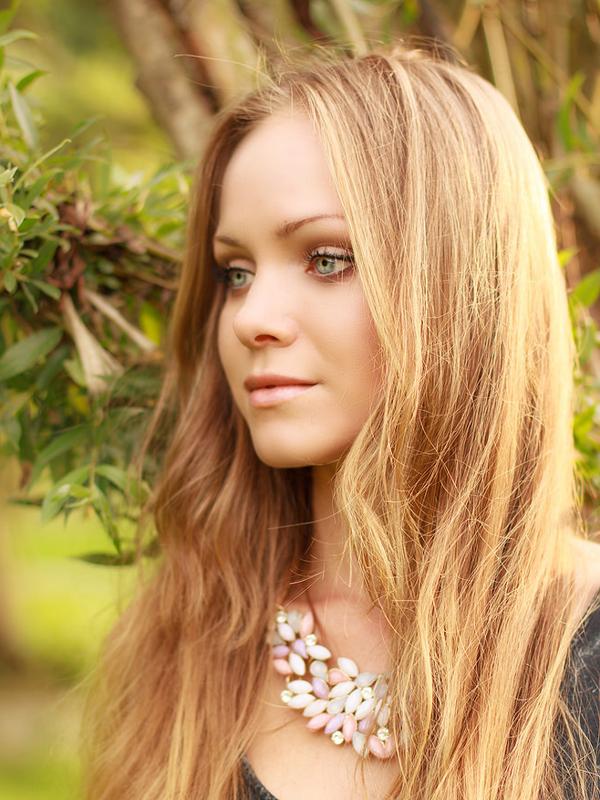}
\end{minipage}%
\hspace{0.06cm}
\begin{minipage}{.32\linewidth}
\begin{overpic}[width=1\linewidth]%
{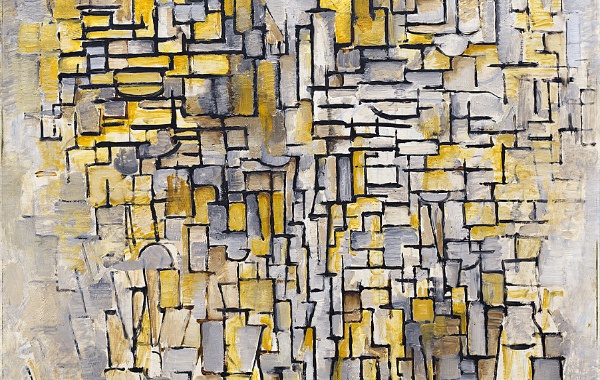}
\put(53.1,0){\includegraphics[width=0.3\linewidth]%
{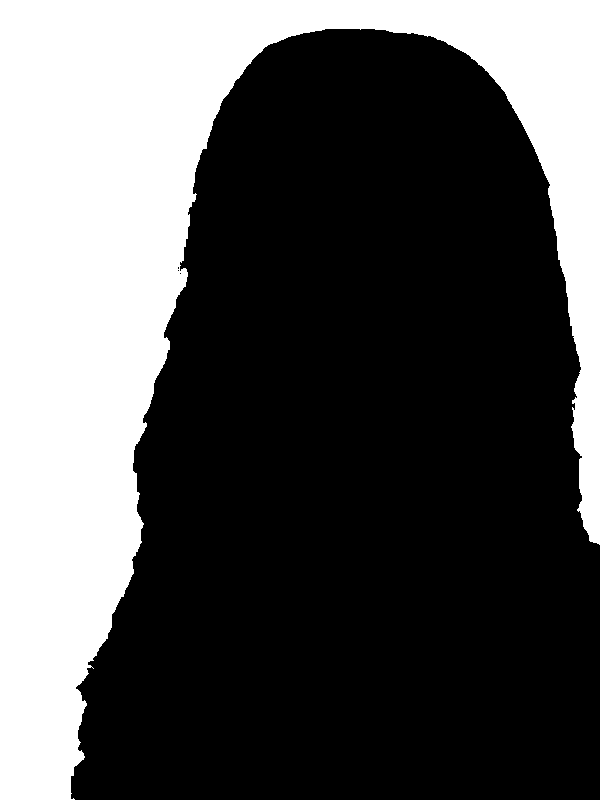}}
\end{overpic}
\vspace{-0.25cm}

\begin{overpic}[width=1\linewidth]%
{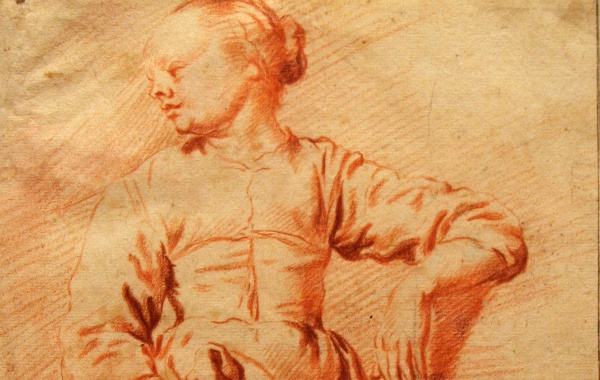}
\put(53.1,0){\includegraphics[width=0.3\linewidth]%
{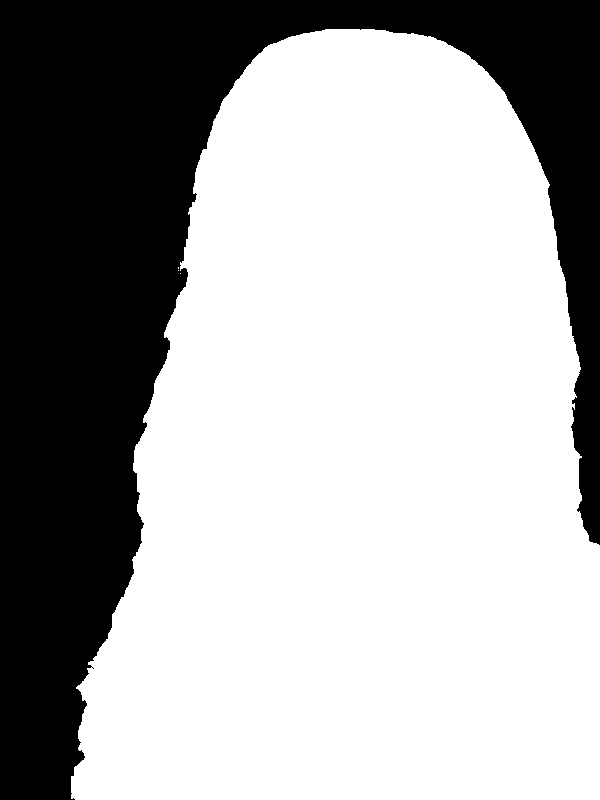}}
\end{overpic}
\end{minipage}
\hspace{0.001cm}
\begin{minipage}{.32\linewidth}
\includegraphics[width=1\linewidth]{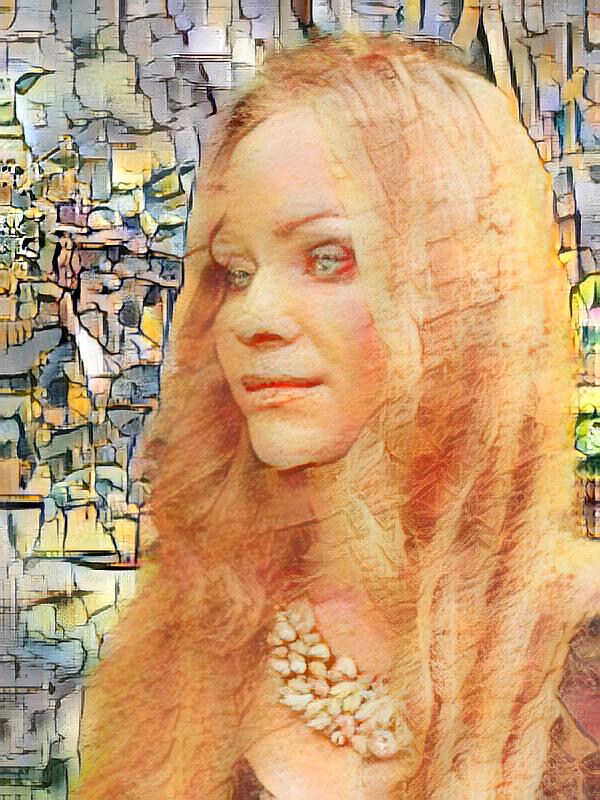}
\end{minipage}

\vspace{0.2cm}
\caption{Spatial control. Left: content image. Middle: two style images with corresponding masks. Right: style transfer result.}
\vspace{-0.1cm}
\label{fig:spatial}
\end{figure}

In Fig.~\ref{fig:spatial} we demonstrate that our method can transfer different regions of the content image to different styles. This is achieved by performing AdaIN separately to different regions in the content feature maps using statistics from different style inputs, similar to~\cite{champandard2016semantic,gatys2017controlling} but in a completely feed-forward manner. 
While our decoder is only trained on inputs with homogeneous styles, it generalizes naturally to inputs in which different regions have different styles.


%



\section{Discussion and Conclusion}
In this paper, we present a simple adaptive instance normalization~(AdaIN) layer that for the first time enables arbitrary style transfer in real-time. 
Beyond the fascinating applications, we believe this work also sheds light on our understanding of deep image representations in general.

It is interesting to consider the conceptual differences between our approach and previous neural style transfer methods based on feature statistics.  Gatys~\etal~\cite{gatys2016image} employ an optimization process to manipulate pixel values to match feature statistics. 
The optimization process is replaced by feed-forward neural networks in~\cite{johnson2016perceptual,ulyanov2016texture,ulyanov2017improved}. Still, the network is trained to modify pixel values to \emph{indirectly} match feature statistics. We adopt a very different approach that \emph{directly} aligns statistics in the feature space \emph{in one shot}, then inverts the features back to the pixel space. 


Given the simplicity of our approach, we believe there is still substantial room for improvement. 
In future works we plan to explore more advanced network architectures such as the residual architecture~\cite{johnson2016perceptual} or an architecture with additional skip connections from the encoder~\cite{pix2pix2017}. We also plan to investigate more complicated training schemes like the incremental training~\cite{li2017diversified}. Moreover, our AdaIN layer only aligns the most basic feature statistics (mean and variance). It is possible that replacing AdaIN with correlation alignment~\cite{sun2016return} or histogram matching~\cite{wilmot2017stable} could further improve quality by transferring higher-order statistics. Another interesting direction is to apply AdaIN to texture synthesis.

\begin{figure*}[!htbp]
\centering
\small

\includegraphics[width=0.162\linewidth]{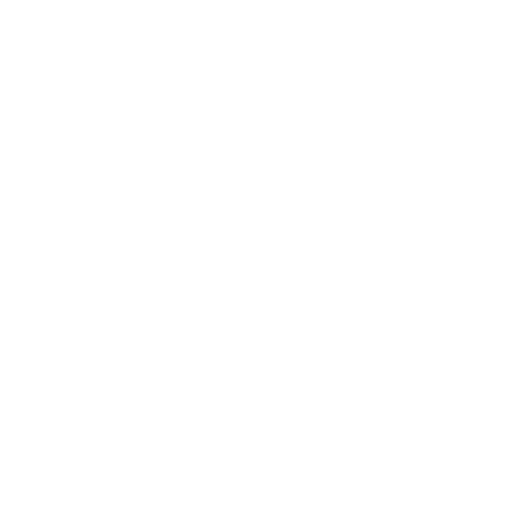}
\includegraphics[width=0.162\linewidth]{figures/cornell1.jpg}
\includegraphics[width=0.162\linewidth]{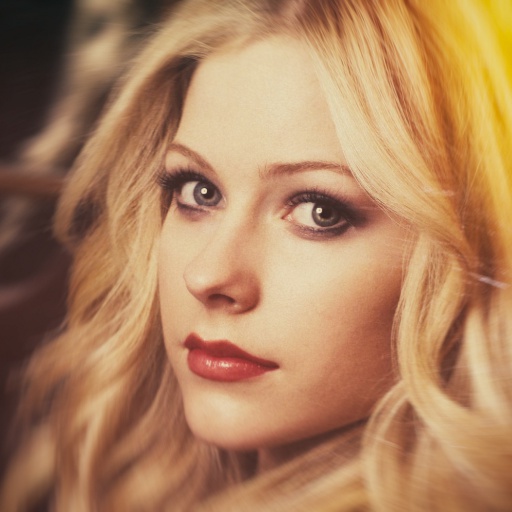}
\includegraphics[width=0.162\linewidth]{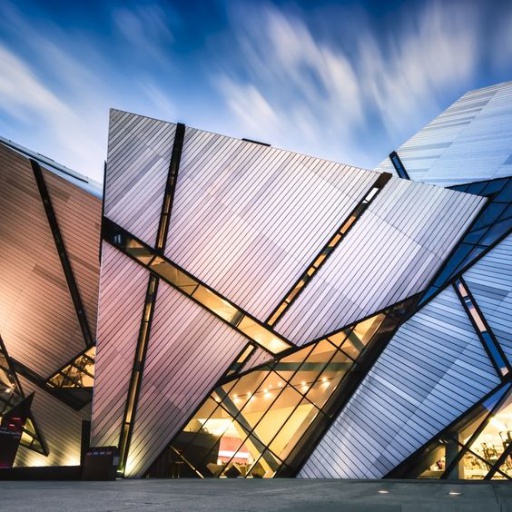}
\includegraphics[width=0.162\linewidth]{figures/lenna.jpg}
\includegraphics[width=0.162\linewidth]{figures/golden_gate.jpg}
\vspace{0.03cm}

\includegraphics[width=0.162\linewidth]{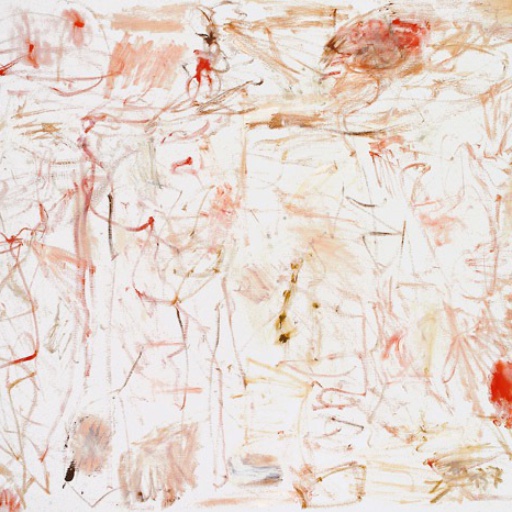}
\includegraphics[width=0.162\linewidth]{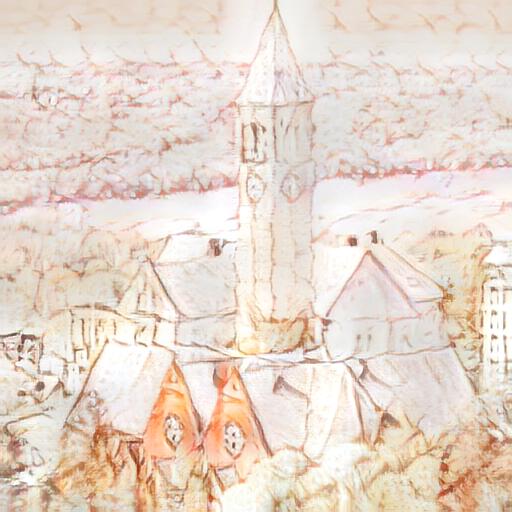}
\includegraphics[width=0.162\linewidth]{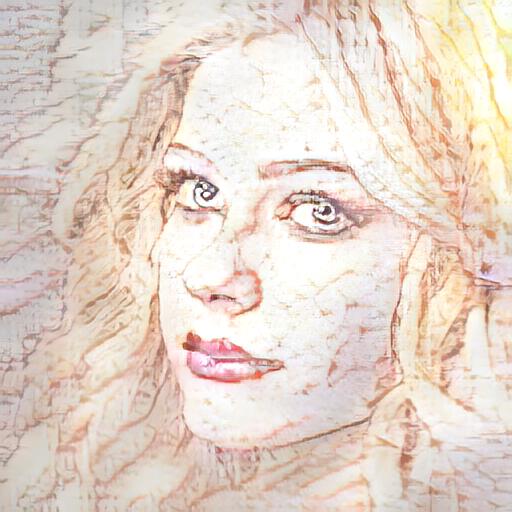}
\includegraphics[width=0.162\linewidth]{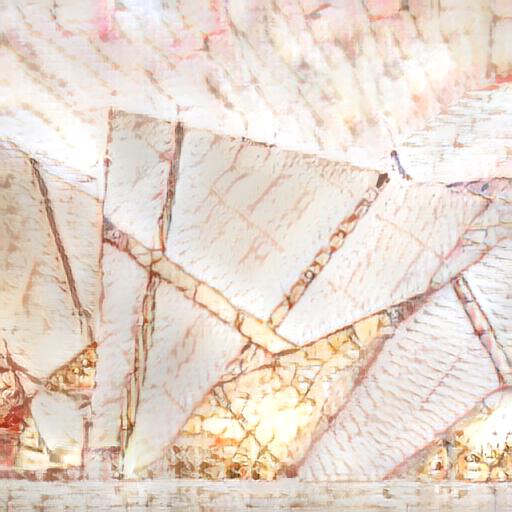}
\includegraphics[width=0.162\linewidth]{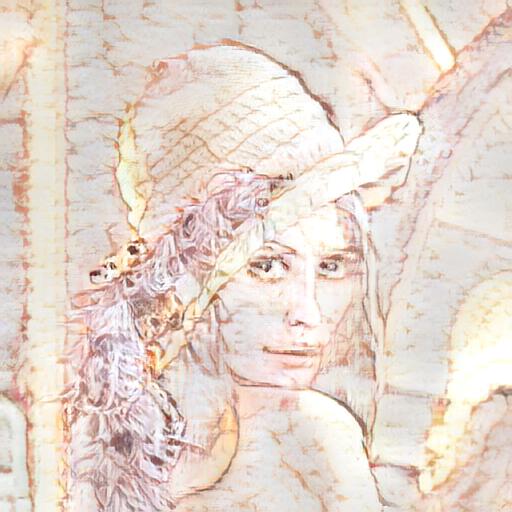}
\includegraphics[width=0.162\linewidth]{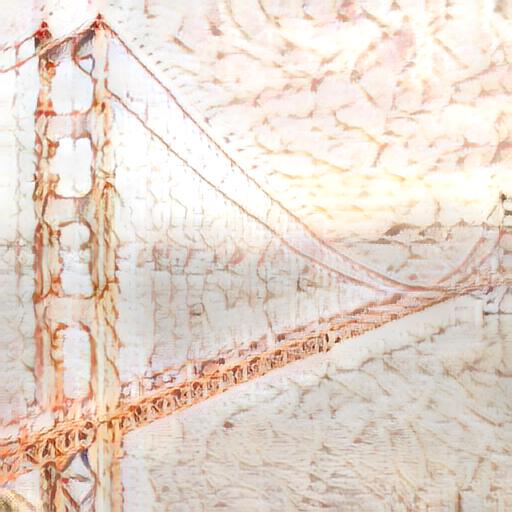}
\vspace{0.03cm}

\includegraphics[width=0.162\linewidth]{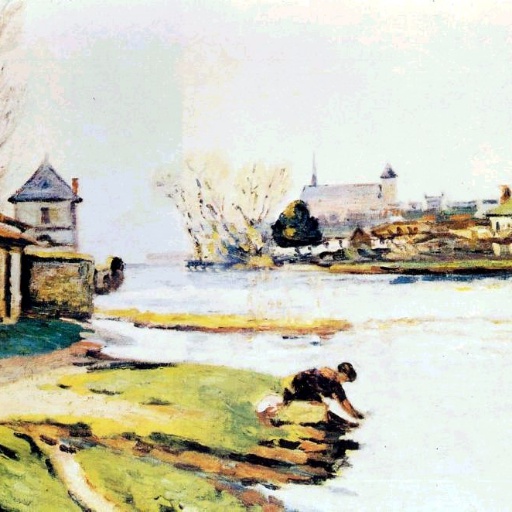}
\includegraphics[width=0.162\linewidth]{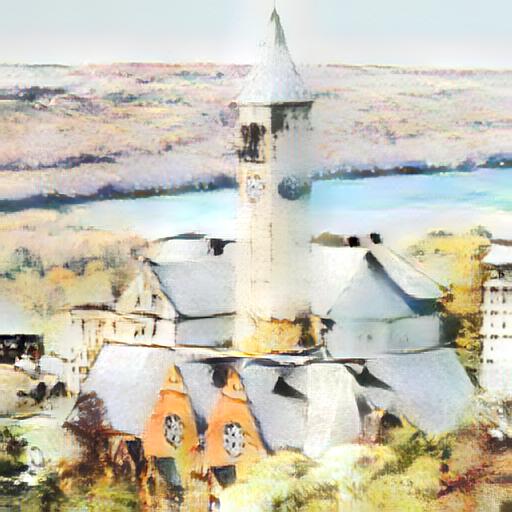}
\includegraphics[width=0.162\linewidth]{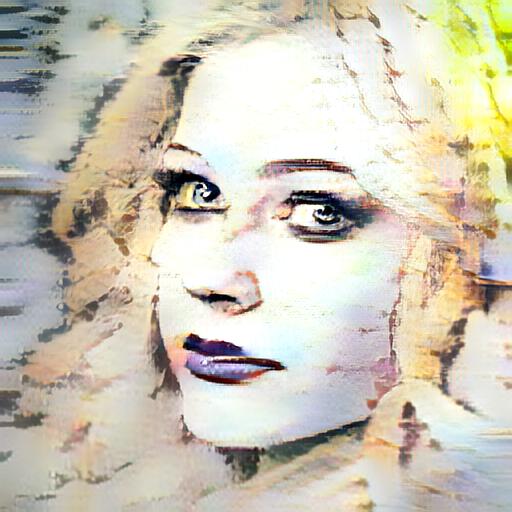}
\includegraphics[width=0.162\linewidth]{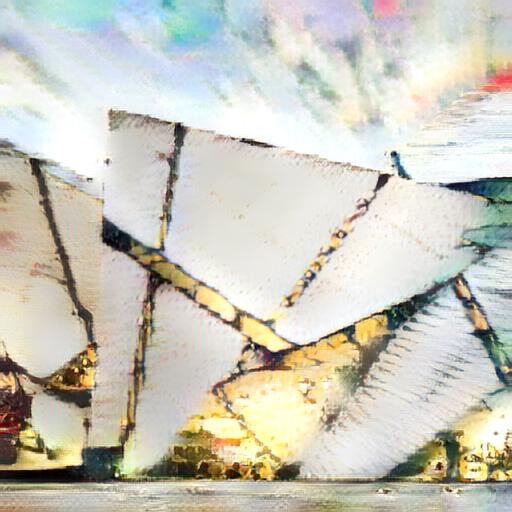}
\includegraphics[width=0.162\linewidth]{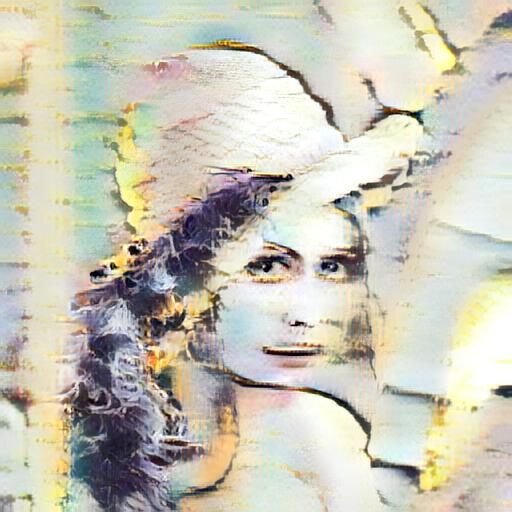}
\includegraphics[width=0.162\linewidth]{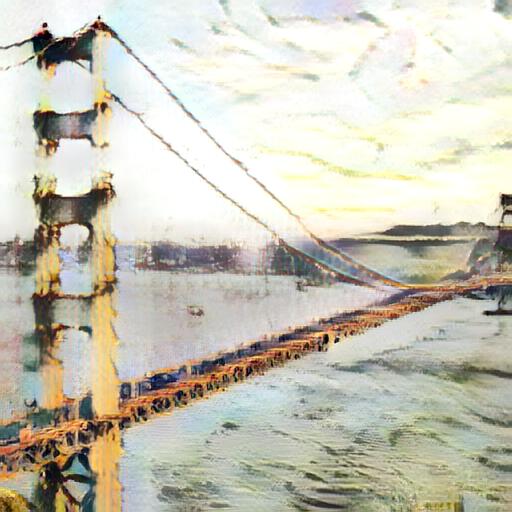}
\vspace{0.03cm}

\includegraphics[width=0.162\linewidth]{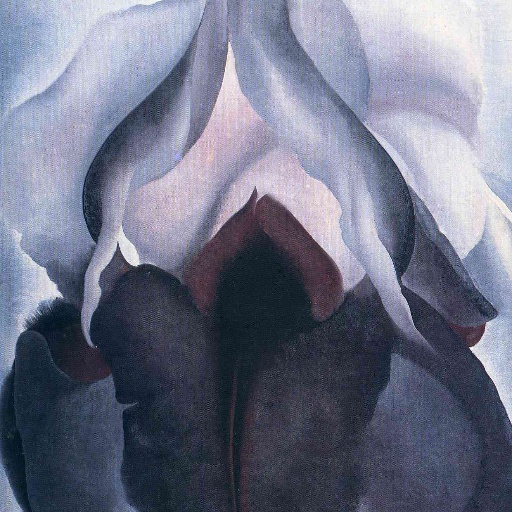}
\includegraphics[width=0.162\linewidth]{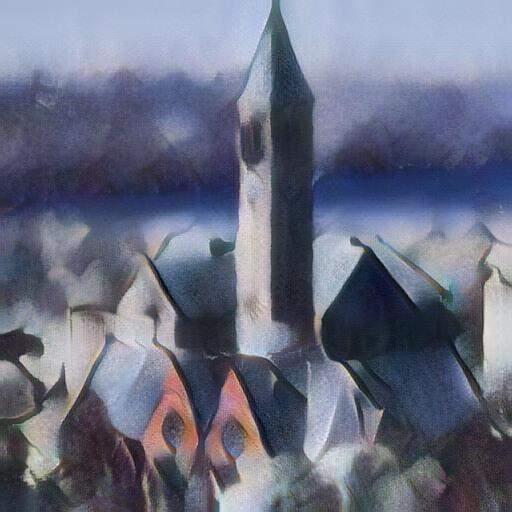}
\includegraphics[width=0.162\linewidth]{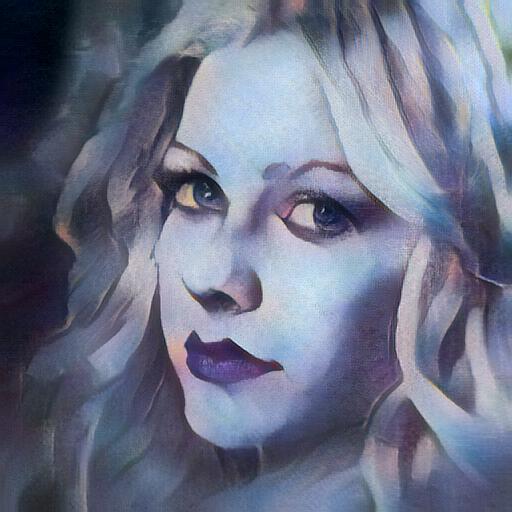}
\includegraphics[width=0.162\linewidth]{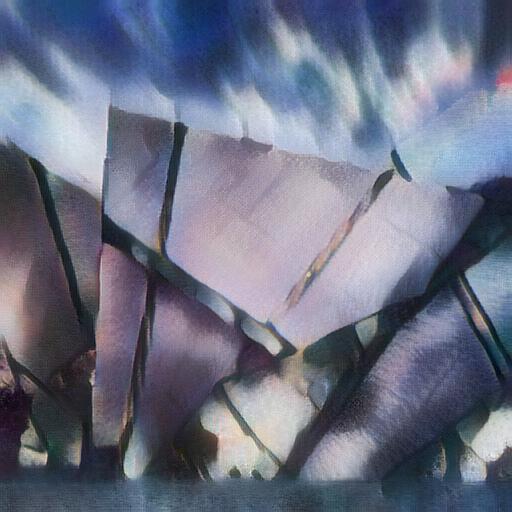}
\includegraphics[width=0.162\linewidth]{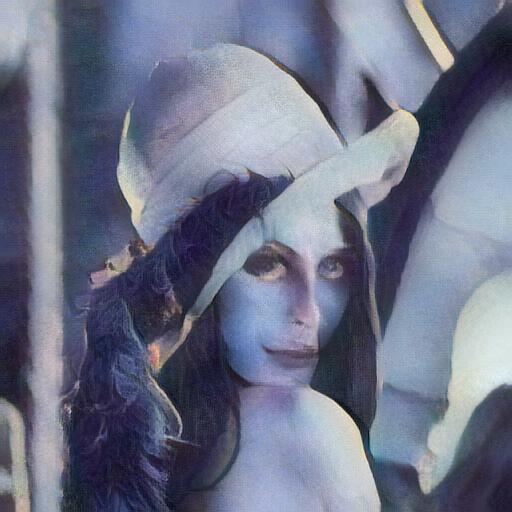}
\includegraphics[width=0.162\linewidth]{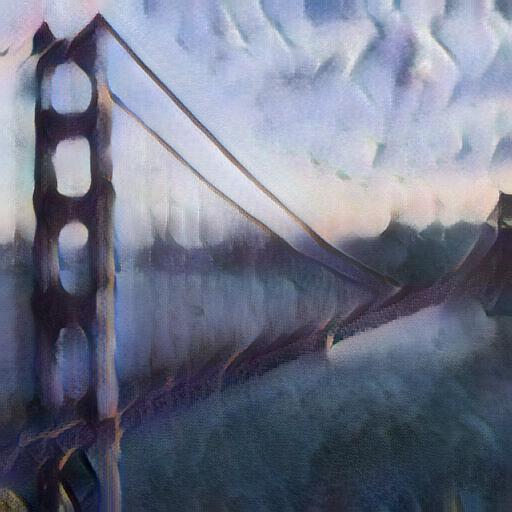}
\vspace{0.03cm}

\includegraphics[width=0.162\linewidth]{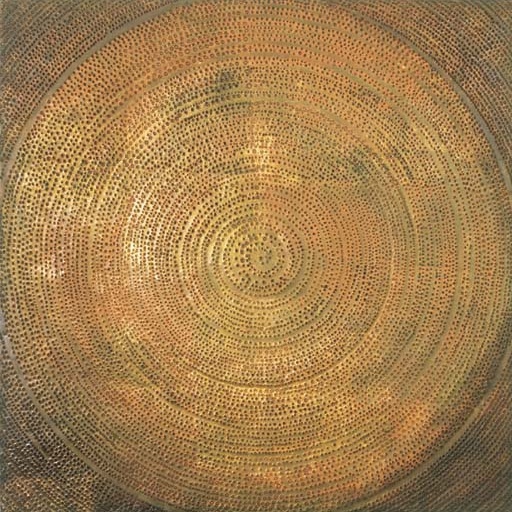}
\includegraphics[width=0.162\linewidth]{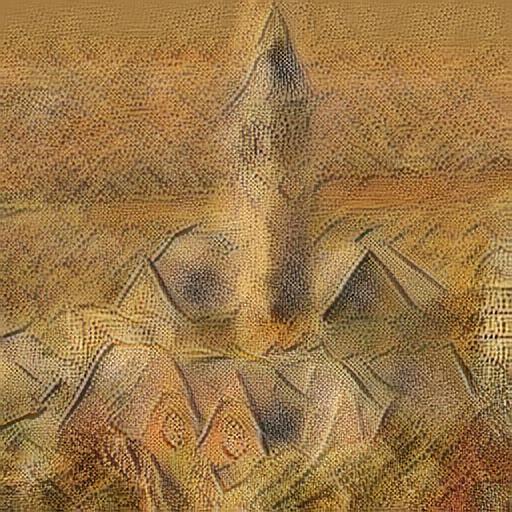}
\includegraphics[width=0.162\linewidth]{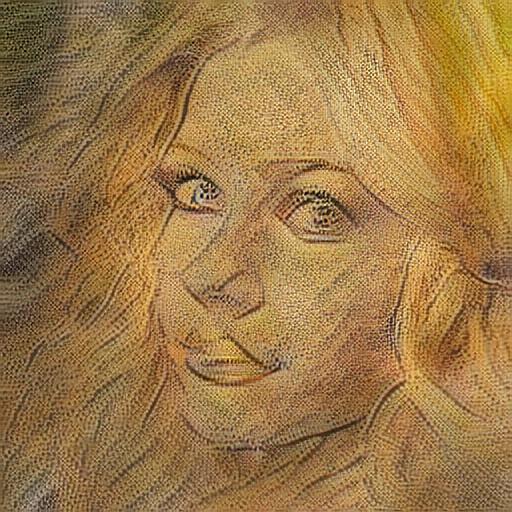}
\includegraphics[width=0.162\linewidth]{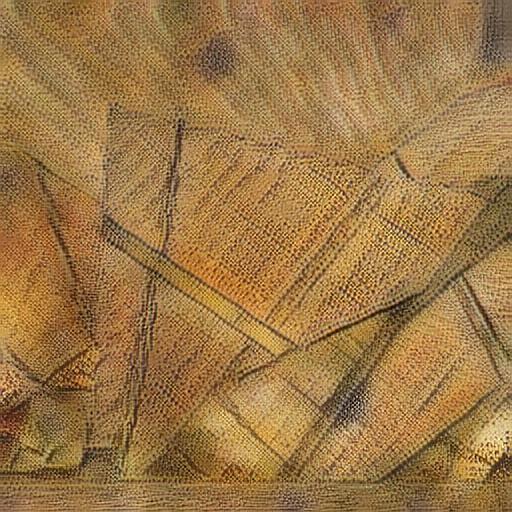}
\includegraphics[width=0.162\linewidth]{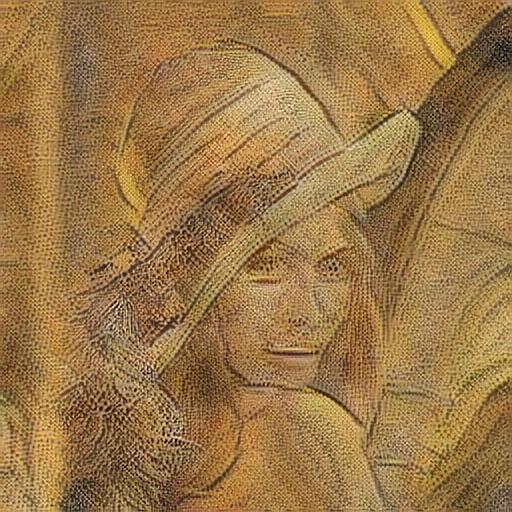}
\includegraphics[width=0.162\linewidth]{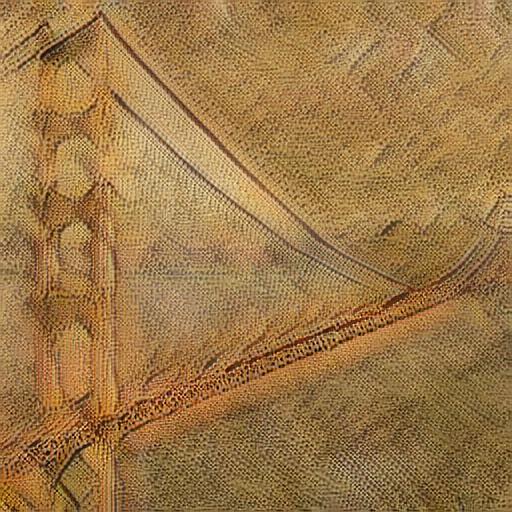}
\vspace{0.03cm}

\includegraphics[width=0.162\linewidth]{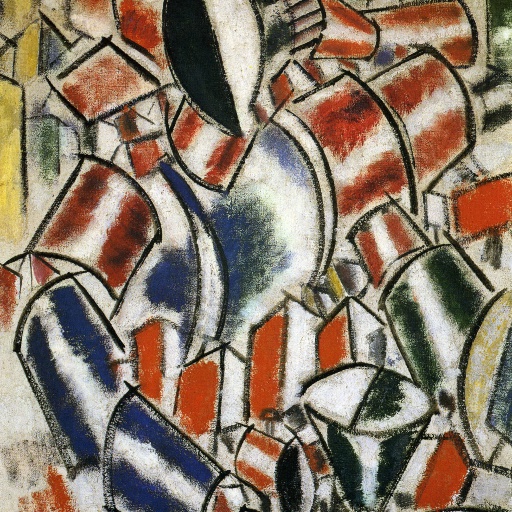}
\includegraphics[width=0.162\linewidth]{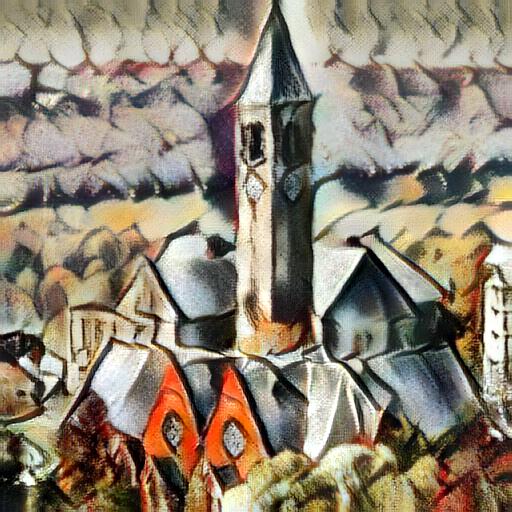}
\includegraphics[width=0.162\linewidth]{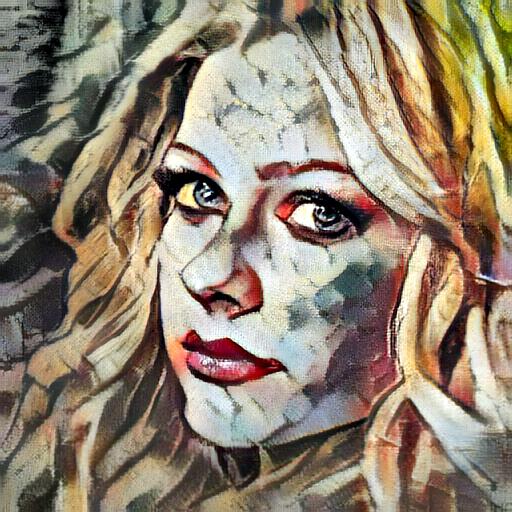}
\includegraphics[width=0.162\linewidth]{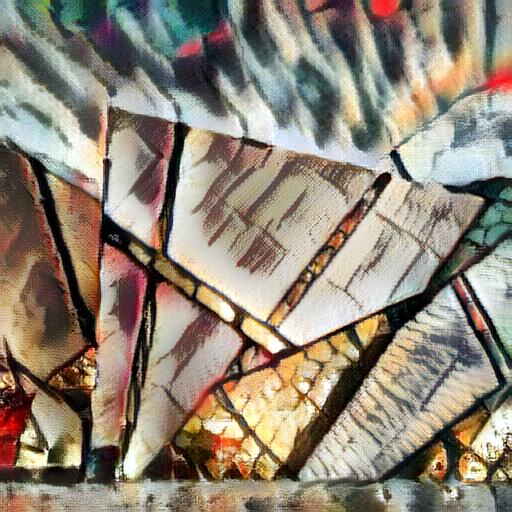}
\includegraphics[width=0.162\linewidth]{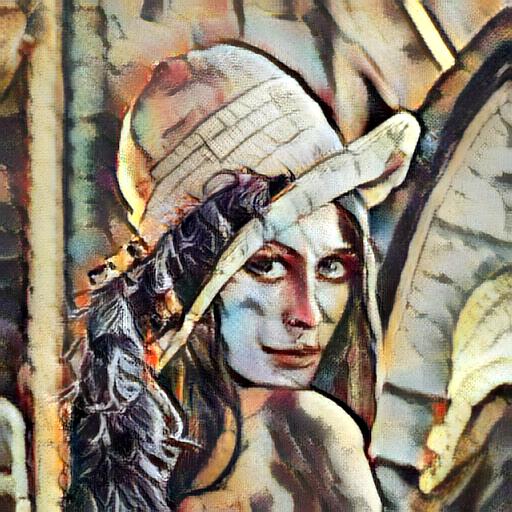}
\includegraphics[width=0.162\linewidth]{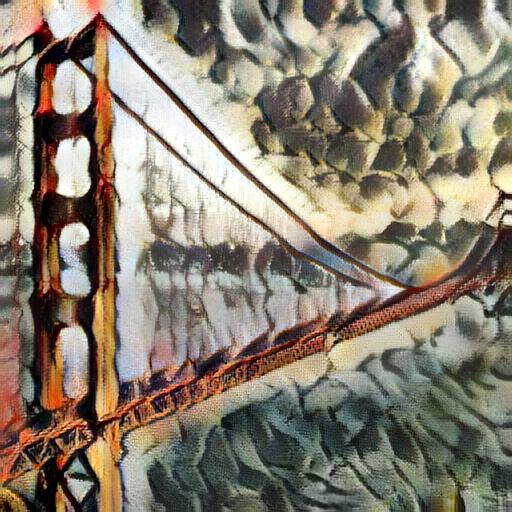}
\vspace{0.03cm}

\includegraphics[width=0.162\linewidth]{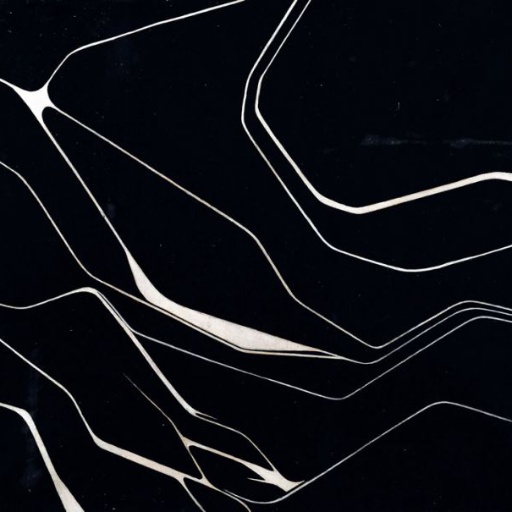}
\includegraphics[width=0.162\linewidth]{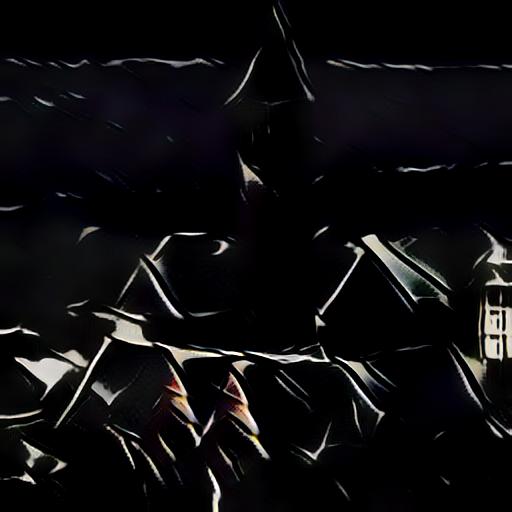}
\includegraphics[width=0.162\linewidth]{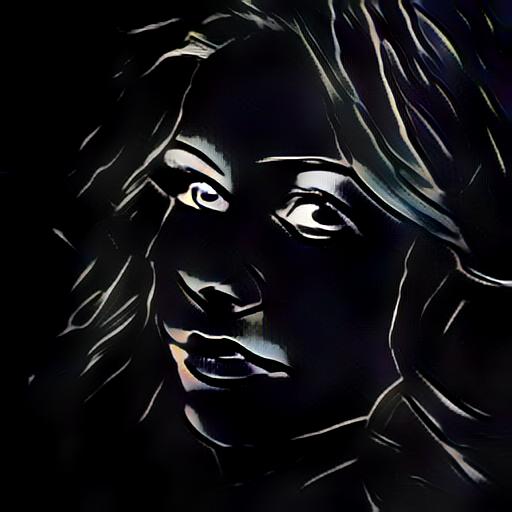}
\includegraphics[width=0.162\linewidth]{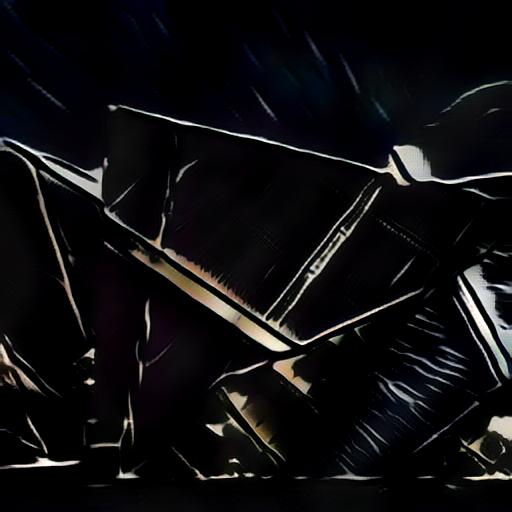}
\includegraphics[width=0.162\linewidth]{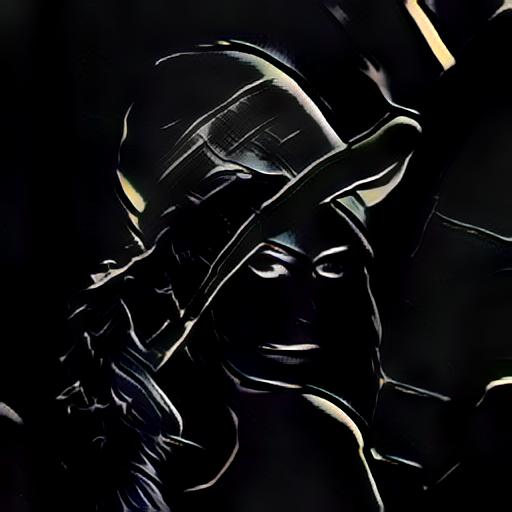}
\includegraphics[width=0.162\linewidth]{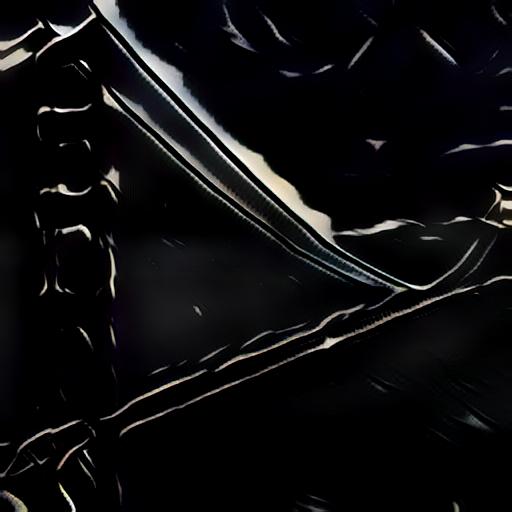}
\vspace{0.23cm}
\caption{More examples of style transfer. Each row shares the same style while each column represents the same content. As before, the network has never seen the test style and content images.}

\vspace{0.3cm}
\label{fig:moreexamples}
\end{figure*}
\vspace{-0.4cm}
\subsubsection*{Acknowledgments}
\vspace{-0.2cm}
We would like to thank Andreas Veit for helpful discussions. This work was supported in part by a Google Focused Research Award, AWS Cloud Credits for Research and a Facebook equipment donation.

{\small
\bibliographystyle{ieee}
\bibliography{egbib}
}

\end{document}